%% file: emnlp2023.tex
\pdfoutput=1

\documentclass[11pt]{article}

\usepackage{EMNLP2023}

\usepackage{times}
\usepackage{latexsym}

\usepackage[T1]{fontenc}

\usepackage[utf8]{inputenc}
\usepackage{enumitem}
\usepackage{microtype}
\usepackage{cleveref}
\usepackage{xcolor}
\usepackage{multirow}
\usepackage{adjustbox}
\usepackage{booktabs}
\usepackage{inconsolata}

\title{Retrieving Multimodal Information for Augmented Generation: A Survey}

\newcommand*{\affaddr}[1]{#1} 
\newcommand*{\affmark}[1][*]{\textsuperscript{#1}}
\newcommand*{\email}[1]{\textrm{#1}}

\author{
  Ruochen Zhao\affmark[1]~~
  Hailin Chen\affmark[1]~~
  \textbf{Weishi Wang\affmark[1]}~~
  Fangkai Jiao\affmark[1]~~ \\
  \textbf{Xuan Long Do}\affmark[1]\thanks{~~Now affiliated with NUS}~~
  \textbf{Chengwei Qin\affmark[1]}~~
  \textbf{Bosheng Ding\affmark[1]}~~
  \textbf{Xiaobao Guo\affmark[1]}~~ \\
  \textbf{Minzhi Li \affmark[2]}~~
  \textbf{Xingxuan Li \affmark[1]}~~
  \textbf{Shafiq Joty\affmark[1,3]\thanks{~~Work done while the author is on leave from NTU}}~~ \\
\affaddr{\affmark[1]Nanyang Technological University, Singapore} \\
\affaddr{\affmark[2]National University of Singapore, Singapore} \\
\affaddr{\affmark[3]Salesforce Research} \\
\email{\small{\{ruochen002, hailin001, xuanlong001, weishi001, fangkai002, chengwei003, bosheng001\}@e.ntu.edu.sg}}\\
\email{\small{\{xiaobao001, xingxuan001\}@e.ntu.edu.sg, li.minzhi@u.nus.edu, srjoty@ntu.edu.sg}}\\
}

\begin{document}
\maketitle
\begin{abstract}
As Large Language Models (LLMs) become popular, there emerged an important trend of using multimodality to augment the LLMs' generation ability, which enables LLMs to better interact with the world.
However, there lacks a unified perception of at which stage and how to incorporate different modalities.
In this survey, we review methods that assist and augment generative models by retrieving multimodal knowledge, whose formats range from images, codes, tables, graphs, to audio.
Such methods offer a promising solution to important concerns such as factuality, reasoning, interpretability, and robustness. 
By providing an in-depth review, this survey is expected to provide scholars with a deeper understanding of the methods' applications and encourage them to adapt existing techniques to the fast-growing field of LLMs.
\end{abstract}

\vspace{-1ex}
\section{Introduction}
\vspace{-1ex}
\input{sections/1_intro}

\vspace{-1ex}
\section{Definitions and Background}
\vspace{-1ex}
\input{sections/2_background}
\vspace{-1ex}
\section{Multimodal Retrieval-Augmented Generation}
\vspace{-1ex}

For each modality, there are different retrieval and synthesis procedures, targeted tasks, and challenges. Therefore, we discuss relevant methods by grouping them in terms of modality, including image, code, structured knowledge, audio, and video.


\vspace{-1ex}
\subsection{Image}
\input{sections/3_image}

\vspace{-1ex}
\subsection{Code}
\input{sections/3_code}

\vspace{-1ex}
\subsection{Structured Knowledge}
\vspace{-1ex}
\input{sections/3_structured}

\vspace{-1ex}
\subsection{Audio}
\vspace{-1ex}
\input{sections/3_audio}

\vspace{-1ex}
\subsection{Video}
\vspace{-1ex}
\input{sections/3_video}

\vspace{-1ex}
\section{Future Directions}
\vspace{-1ex}
\input{sections/4_future}

\vspace{-1ex}
\section{Conclusions}
\vspace{-1ex}
\input{sections/5_conclusion}

\section*{Limitations}

RAG also has some limitations. For example, there exists an attribution-fluency tradeoff \citep{aksitov2023characterizing} where the output quality is affected due to the added constraints of the retrieved knowledge.

\section*{Acknowledgement}
This research is supported by the National Research Foundation, Singapore under its AI Singapore Programme (AISG Award No: AISG-PhD/2021-01-001). This research is supported, in part, by Alibaba-NTU Singapore Joint Research Institute (JRI), Nanyang Technological University, Singapore.

\bibliography{anthology,custom}
\bibliographystyle{acl_natbib}

\appendix

\clearpage
\section{Appendix}
\label{sec:appendix}
\input{sections/app}


\end{document}

%% file: sections/1_intro.tex

Generative Artificial Intelligence (GAI) has demonstrated impressive performances in tasks such as text generation \citep{ouyang2022training, chowdhery2022palm, brown2020language} and text-to-image generation \citep{DALL-E, poole2022dreamfusion}. The recent advancements in Multimodal Large Language Models (MLLMs) \citep{driess2023palm, gpt4, huang2023language} have further improved the models' capabilities to handle multi-format information, opening up possibilities for developing general-purpose learners. 

Nevertheless, generative models are not exempt from inherent limitations, including the tendency for generating hallucinations~\citep{ye2022unreliability}, struggling with arithmetic tasks~\citep{math}, and lacking interpretability
. Consequently, a promising solution for enhancing their capabilities lies in enabling them to interact with the external world and acquire knowledge in diverse formats and modalities, thereby improving the factuality and rationality of the generated content.
Recently, there have been emerging studies focusing on retrieval-augmented approaches \citep{mialon2023augmented}, which aim to provide information that is more grounded and factually dependent. Among them, most \citep{nakano2021webgpt, guu2020retrieval} retrieves textual information, 
which matches the data format used during pre-training and offers a natural medium for interaction. However, there is more world knowledge stored in different structures and modalities, such as images and videos, which is often inaccessible, unavailable, or not describable in traditional textual corpora.

Therefore, there arises an important research intersection that retrieves multimodal knowledge to augment generative models. It offers a promising solution to current challenges such as factuality, reasoning, interpretability, and robustness. As this field is very recent, there lacks a unified understanding of recognizing these methods as a specific group, visualizing their innate connections, connecting their methodologies, and outlining their applications.


Therefore, we survey recent advancements in multimodal retrieval-augmented generation (RAG). Specifically, we discuss current research by grouping them into different modalities, including image, code, structured knowledge, audio, and video.
For each modality, we systematically search the ACL Anthology and Google Scholar with relevant keywords and perform manual filtering to determine their relevance to the survey. As a result, we collect 146 papers for detailed analysis. 
In \Cref{subsec:search}, we include search details, statistics, and a trend analysis figure, which shows that multimodal RAG papers have indeed developed very fastly since the emergence of large-scale general-purpose models. Within each modality, we discuss relevant papers by grouping them under different applications. By providing an in-depth survey, we hope to help researchers recognize the importance of incorporating knowledge in different formats and encourage adaptation and advancements on existing techniques to the fast-growing field of LLMs.

In summary, our contributions are as follows:

\vspace{-0.5ex}
\begin{itemize}[leftmargin=*,topsep=2pt,itemsep=2pt,parsep=0pt]
\item We establish retrieval augmented generation with multi-modality as an important group of methods that emerges with the recent advances in LLMs.
\item For common modalities, we provide an in-depth review of research papers by contextualizing their innate connections and shared challenges.
\item We provide an informative analysis of future directions, which could contain promising solutions to many current challenges.
\end{itemize}
\vspace{-0.5ex}




%% file: sections/2_background.tex
To better understand the state and advancements that inspired multimodal retrieval augmentation, we first define and discuss the background of two key concepts: multimodal learning and retrieval-augmented generation (RAG).

\vspace{-1ex}
\subsection{Multimodal Learning} 

Multimodal learning refers to learning a unified representation of data from different modalities. It aims at extracting complementary information to facilitate compositional tasks~\cite{baltruvsaitis2018multimodal, gao2020survey}. {In this survey, we include all modalities whose formats are different from natural language, including image, code, structured knowledge (\emph{e.g.} tables, knowledge graphs), audio, and video.}

Multimodal generative models have a wide range of applications, such as text-image generation, creative writing generation, and multilingual translation. For instance, the image recognition task can benefit from analyzing images and videos in conjunction with textual descriptions~\cite{ju2022prompting, alayracflamingo, jia2021scaling, radford2021learning}. Conversely, incorporating visual information
also aids language understanding and generation~\citep{zhou2020unified, lei2021less, otter}. Moreover, they have the potential to significantly improve machine learning systems across various domains by enabling models to learn from and integrate multiple sources of information~\cite{tsai2019multimodal, acosta2022multimodal, nagrani2021attention}.
Additionally, there has been growing interest in developing generative models that can output multiple modalities of data \citep{ramesh2021zero, crowson2022vqgan, RA-VQA, chen2022murag}. However, there remain challenges for multimodal generative models, such as gaining access to a large amount of multimodal data and designing a network that produces semantically meaningful outputs.

\vspace{-1ex}
\subsection{Retrieval-Augmented Generation (RAG)}

RAG 
typically consists of two phases: 
retrieving contextually relevant information, and guiding the generation process using the retrieved knowledge.

Recently, RAG has gained popularity in Natural Language Processing (NLP) due to the rise of general-purpose Large Language Models (LLMs) \cite{chowdhery2022palm, gpt4}, which have boosted performances in a wide range of NLP tasks. However, there are two primary challenges: Firstly, because generative models rely on the inner knowledge (weights), they result in a high amount of hallucinations \cite{zhao2023chatgptlike}.
Secondly, due to their large parameter sizes and the high costs of updating, the traditional pretraining and finetuning approaches have become infeasible. As a solution, RAG methods  \cite{gu2018search,weston-etal-2018-retrieve,cai-etal-2019-retrieval,lewis2020retrieval} offer a promising solution for LLMs to effectively interact with the external world.

RAG is applied to a wide range of downstream NLP tasks, including machine translation~\cite{gu2018search, zhang-etal-2018-guiding, xu2020boosting, he2021fast}, dialogue generation \cite{weston-etal-2018-retrieve, wu2019response,cai-etal-2019-skeleton}, abstractive summarization \cite{peng-etal-2019-text}, and knowledge-intensive generation \cite{lewis2020retrieval,izacard-grave-2021-leveraging}. Among them, most methods focus on retrieving textual information.
For example, \citet{guu2020retrieval, lewis2020retrieval, borgeaud2022improving, izacard2022atlas} jointly train a retrieval system with an encoder or sequence-to-sequence LM, achieving comparable performance to larger LMs that use significantly more parameters. 
Recent research also proposes combining a retriever with chain-of-thought (CoT) prompting for reasoning to augment language models \cite{he2022rethinking,trivedi2022interleaving,zhao2023verifyandedit}. 

%% file: sections/3_image.tex

 
Recent advances on pretrained models shed light on general image-text multi-modal models \citep{DALL-E, flamingo, CM3, parti, dou2022coarse, li2023blip}. However, these models require huge computational resources for pretraining and large amounts of model parameters --- as they need to memorize vast world knowledge.
More critically, they cannot efficiently deal with new or out-of-domain knowledge. 
To this end, multiple retrieval-augmented methods have been proposed to better incorporate external knowledge from images and text documents. 
In general text generation tasks, image retrieval can also improve generation quality by expanding text generation contexts with more ``imagination''.

\noindent\textbf{Visual question answering (VQA)}~~~
To tackle open-domain VQA, RA-VQA \cite{lin-byrne-2022-retrieval} jointly trains the document retriever and answer generation module by approximately marginalizing predictions over retrieved documents. 
It first uses existing tools of object detection, image captioning, and optical character recognition (OCR) to convert target images to textual data. Then, it performs dense passage retrieval (DPR)~\citep{dpr} to fetch text documents relevant to the target image in the database. Finally, each retrieved document is concatenated with the initial question to generate the final prediction, similar to RAG \cite{lewis2020retrieval}. Besides external documents, PICa \citep{yang2022empirical} and KAT \cite{gui-etal-2022-kat} also consider LLMs as implicit knowledge bases and extract relevant implicit information from GPT-3. Plug-and-Play \cite{tiong-etal-2022-plug} retrieves relevant image patches by using GradCAM \cite{GradCAM} to localize relevant parts based on the initial question. It then performs image captioning on retrieved patches to acquire augmented context. Beyond text-only augmented context, MuRAG \cite{chen-etal-2022-murag} retrieves both text and image data and incorporates images as visual tokens. RAMM \cite{RAMM} retrieves similar biomedical images and captions and encodes them through different networks. 

\noindent\textbf{Image captioning}~~~
To generate multi-style captions, \citet{zhou-long-2023-style} uses a style-aware visual encoder to retrieve image contents before generating captions. Beyond simply encoding visual information, \citet{cho-etal-2022-fine} further uses the multimodal similarity between  image-text pairs as a reward function to train a more fine-grained captioning model. Beyond retrieving image elements, \citet{RA-transformer, shi-etal-2021-retrieval-analogy,ramos-etal-2023-retrieval,Re-ViLM} retrieves relevant captions to the inputs. \citet{zhou-etal-2022-focus} tackles news image captioning by retrieving visually grounded entities in news articles.

\noindent\textbf{Visually grounded dialogue}~~~
This task \citep{lee-etal-2021-constructing} requires retrieving visual information to produce relevant dialog responses. \citet{fan-etal-2021-augmenting} augments generative models with KNN-based Information Fetching (KIF) modules that retrieve images and wiki knowledge. \citet{liang-etal-2021-maria} retrieves a correlated image to the dialog from an image index to ground the response generator. \citet{text_is_not_enough} trains a word-image mapping model to retrieve response visual impressions and then uses both textual and visual information for response generation.

\noindent\textbf{Text generation}~~~
For general text generation tasks, image retrieval can also help expand contexts. \citet{yang-etal-2022-z} augments a text model's ``imagination '' by retrieving existing images and synthesizing newly generated images. As a result, fueling language models with imagination can improve performances in many downstream natural language tasks. Similarly, \citet{zhu-etal-2023-visualize} compares ``imagination'' augmentation with synthetic and retrieved images and argues that machine-generated images could provide better guidance due to better consideration of the contexts. Moreover, \citet{fang-feng-2022-neural} shows that machine translation can be significantly improved by retrieving visual information at the phrase level, especially when the textual context is limited. Image RAG can also help low-resource tasks such as medical report generation \citep{wu-etal-2022-deltanet} and architectural description generation \citep{mille-etal-2020-case}.

Beyond retrieving images before generating text, Re-Imagen \citep{chen2022re} leverages a  multi-modal knowledge base to retrieve image-text pairs to facilitate image generation. RA-CM3 \cite{RA-CM3} can generate mixtures of images and text. It shows that retrieval-augmented image generation performs much better on knowledge-intensive generation tasks and opens up new capabilities such as multimodal in-context learning. 

%% file: sections/3_code.tex
Software developers attempt to search for relevant information to improve their productivity from large amounts of available resources such as explanations for unknown terminologies, reusable code patches, and solutions to common programming bugs~\cite{dsf}. Inspired by the progress of deep learning in NLP, a general retrieval-augmented generation paradigm has benefited a wide range of code intelligent tasks, including code completion \cite{lu-etal-2022-reacc}, code generation \cite{zhou2022docprompting}, and automatic program repair (APR) \cite{nashidretrieval}. However, these approaches often treat programming languages and natural languages as equivalent sequences of tokens and ignore the rich semantics inherent to source code. To address these limitations, recent research work has focused on improving code generalization performance via multimodal learning, which incorporates additional modalities such as code comments, identifier tags, and abstract syntax trees (AST) into code pretrained models~\cite{wang-etal-2021-codet5,unixcoder,codereviewer}. To this end, multimodal retrieval-augmented generation approach has demonstrated its feasibility in a variety of code-specific tasks.

\noindent\textbf{Text-to-Code Generation}~~~ 
Numerous research studies have investigated the utilization of relevant codes and associated documents to benefit code generation models. A prominent example is REDCODER \cite{parvez-etal-2021-retrieval-augmented}, which retrieves the top-ranked code snippets or summaries from an existing codebase, and aggregates them with source code sequences to enhance the generation or summarization capabilities. As another such approach, DocPrompting \citep{zhou2022docprompting} uses a set of relevant documentation as in-context prompts to generate corresponding code via retrieval. In addition to these lexical modalities, \citet{hayati-etal-2018-retrieval} proposes a syntax-based code generation approach to reference existing subtree from the AST as templates to direct code generation explicitly.

\noindent\textbf{Code-to-Text Generation}~~~
Retrieval-based code summarization methods are studied extensively. For example, RACE~\cite{shi-etal-2022-race} leverages relevant code differences and their associated commit messages to enhance commit message generation. Besides, RACE calculates the semantic similarity between source code differences and retrieved ones to weigh the importance of different input modalities. 
Rencos~\cite{10.1145/3377811.3380383} retrieves two similar code snippets based on the aspects of syntactic-level similarity and semantic-level similarity of a given query code. 
These similar contexts are then incorporated into the summarization model during the decoding phase.
This idea is further explored by~\citet{liu2021retrievalaugmented}, where retrieved code-summary pairs are used to augment the original code property graph~\cite{DBLP:conf/sp/YamaguchiGAR14} of source code via local attention mechanisms. 
To capture the global semantics for better code structural learning, a global structure-aware self-attention mechanism~\cite{zhu-etal-2019-modeling} is further employed.

\noindent\textbf{Code Completion}~
Recent advances in retrieval-based code completion tasks~\cite{mcconnell2004code} have gained increasing attention. Notably,~\citet{hashimoto2018retrieve} adapts the retrieve-and-edit framework to improve the model's performance in code auto-completion tasks. To address practical code completion scenarios, ReACC~\cite{lu-etal-2022-reacc} takes both lexical and semantic information of the unfinished code snippet into account, utilizing a hybrid technique to combine a lexical-based sparse retriever and a semantic-based dense retriever. First, the hybrid retriever searches for a relevant code from the codebase based on the given incomplete code. Then, the unfinished code is concatenated with the retrieval, and an auto-regressive code completion generator will generate the completed code based on them. In order to address project relations, CoCoMIC~\cite{ding2022cocomic} decomposes a code file into four components: \emph{files}, \emph{global variables}, \emph{classes}, and \emph{functions}. It constructs an in-file context graph based on the hierarchical relations among all associated code components, forming a project-level context graph by considering both in-file and cross-file dependencies. Given an incomplete program, CoCoMIC retrieves the most relevant cross-file entities from its project-level context graph and jointly learns the incomplete program and the retrieved cross-file context for code completion.   


\noindent\textbf{Automatic Program Repair (APR)}~~~
Inspired by the nature that a remarkable portion of commits is comprised of existing code commits~\cite{Martinez2014DoTF}, APR is typically treated as a search problem by traversing the search space of repair ingredients to identify a correct fix~\cite{DBLP:conf/icse/QiMLDW14}, based on a redundancy assumption~\cite{White2019SortingAT} that the target fix can often be reconstructed in the search space. Recent studies have shown that mining relevant bug-fix patterns from existing search space~\cite{simfix} and external repair templates from StackOverflow~\cite{DBLP:conf/wcre/LiuZ18} can significantly benefit APR models.~\citet{joshi2022repair} intuitively ranks a collection of bug-fix pairs based on the similarity of error messages to develop few-shot prompts. They incorporate compiler error messages into a large programming language model Codex~\cite{codex} for multilingual APR. CEDAR~\cite{nashidretrieval} further extends this idea to retrieval-based prompts design using relevant code demonstrations, comprising more modalities such as unit test, error type, and error information. Additionally, ~\citet{jin2023inferfix} leverage a static analyzer Infer to extract error type, error location, and syntax hierarchies~\cite{clement2021long} to prioritize the focal context. Then, they retrieve semantically-similar fixes from an existing bug-fix codebase and concatenate the retrieved fixes and focal context to form the instruction prompts for program repair.

\noindent\textbf{Reasoning over Codes as Intermediate Steps}
While large language models (LLMs) have recently demonstrated their impressive capability to perform reasoning tasks, they are still prone to logical and arithmetic errors~\cite{gao2022pal, chen2022program, madaan2022language}. To mitigate this issue, emerging research papers have focused on using LLMs of code (e.g., Codex \cite{codex}) to generate the code commands for solving logical and arithmetic tasks and calling external interpreters to execute the commands to obtain the results. Notably, \citet{gao2022pal} proposes to generate Python programs as intermediate reasoning steps and offload the solution step to a Python interpreter. Additionally, \citet{chen2022program} explore generating chain-of-thought (CoT) \cite{wei2022chain} for not only text but also programming language statements as reasoning steps to solve the problem. During the inference phase, answers are obtained via an external interpreter. Similarly, \citet{lyu2023faithful} propose Faithful CoT that first translates the natural language query to a symbolic reasoning chain and then solves the reasoning chain by calling external executors to derive the answer. Another example is \citet{ye2023large}, which utilizes LLMs to decompose table-based reasoning tasks into subtasks, decouples logic and numerical computations in each step through SQL queries by Codex, and calls SQL interpreters to solve them (a process called "parsing-execution-filling").

LLMs of code are also known as good-structured commonsense reasoners, and even better-structured reasoners than LLMs \cite{madaan2022language}. As a result, prior studies have also investigated the idea of transforming structured commonsense generation tasks into code generation problems and employing LLMs of code as the solvers. One such work is CoCoGen \cite{madaan2022language} which converts each training sample (consisting of textual input and the output structure) into a Tree class in Python. The LLMs of code then perform few-shot reasoning over the textual input to generate Python code, and the Python code is then converted back to the original structure for evaluation. Besides, the success of LLMs of code such as Codex in synthesizing computer code also makes them suitable for generating formal codes. Motivated by this, \citet{wu2022autoformalization} propose to prove mathematical theorems by adopting Codex to generate formalized theorems from natural language mathematics for the interactive theorem prover Isabelle \cite{WenzelPN-TPHOLs08}.


%

%% file: sections/3_structured.tex

An open challenge in generative models is hallucination, where the model is likely to output false information.
Thus, A potential solution is to ground generation with retrieved structured knowledge, such as knowledge graphs, tables, and databases.

\noindent\textbf{Question Answering (QA)}~~~
A natural setting to use knowledge is QA. To augment \emph{Knowledge Base (KB) QA} by extracting the most relevant knowledge, \citet{hu-etal-2022-logical} uses dense retrieval and \citet{liu-etal-2022-uni} uses a cross-encoder ranker. \citet{shu-etal-2022-tiara} employs multi-grained retrieval to extract relevant KB context and uses constrained decoding to control the output space. In \emph{table QA}, \citet{nan-etal-2022-fetaqa} proposes a dataset that requires retrieving relevant tables for answer generation. \citet{pan-etal-2021-cltr} then proposes a method that uses a transformer-based system to retrieve the most relevant tables and locate the correct cells. Moreover, to improve \emph{Video QA}, \citet{hu2023reveal} retrieves from Knowledge Graph (KG) encodings stored in the memory. The most prominent RAG usage remains in \emph{open-domain QA}, where multiple datasets are proposed \citep{lin-etal-2023-fvqa, ramnath-etal-2021-worldly}. For retrieval, \citet{ma-etal-2022-open-domain} verbalizes the KG and then uses dense passage retrieval. \citet{fan2019using, gupta-etal-2018-retrieve} encodes KG information into dense representations. \citet{pramanik2021uniqorn, jin2022heterformer} builds graph embeddings to retrieve question-relevant evidence. \citet{xu-etal-2021-fusing, baek2023knowledge} use semantic similarity and text-matching methods.
{Synthesis can occur at different stages. At the input stage,} \citet{xu-etal-2021-fusing, baek2023knowledge} feed in the retrieved contexts as additional inputs or prompts to the PLM. \citep{ma-etal-2022-open-domain, fan2019using} adapt the generator to accept the context representations as inputs.  {At model inference stage, }an interesting work is \citet{hu-etal-2022-empowering}, which inserts an interaction layer into PLMs to guide an external KG reasoning module.

\noindent\textbf{General text generation}~~~
External knowledge retrieval can improve general text generation to be more factually grounded. \citet{liu-etal-2022-relational} presents a memory-augmented approach to condition an autoregressive language model on a knowledge graph (KG). {During inference, }\citet{tan-etal-2022-tegtok} selects knowledge entries through dense retrieval and then injects them into the input encoding and output decoding stages in pretrained language models (PLMs). For \emph{domain-specific text generation}, \citet{frisoni-etal-2022-bioreader, yang-etal-2021-writing, li2019knowledgedriven} retrieve medical report chunks or report templates {to augment input prompts. Then, they} use self-devised decoders or graph transformers to generate grounded reports. To improve interpretability, RAG could be used to select facts as interpretable reasoning paths \citep{aggarwal-etal-2021-explanations, jansen-ustalov-2019-textgraphs}. 
Moreover, RAG is especially useful for low-resource generation tasks, such as question generation \citep{yu-jiang-2021-expanding, xin-etal-2021-enhancing, gu-etal-2019-extract}, document-to-slide \citep{sun-etal-2021-d2s}, table-to-text \citep{su-etal-2021-shot-table}, counterargument generation \citep{jo-etal-2021-knowledge-enhanced}, entity description generation \citep{cheng-etal-2020-ent} and text-based games \citep{murugesan-etal-2021-efficient}.

Recent research has attempted to reduce hallucinations in LLMs by leveraging external structured knowledge. For example, during fine-tuning, LaMDA \citep{thoppilan2022lamda} learns to consult external knowledge sources before responding to the user, including an information retrieval system that can retrieve knowledge triplets and web URLs. Some papers treat the generative model (often large language models) as black-box and retrieve structured information without fine-tuning. For example, BINDER \citep{cheng2022binding} uses in-context learning to output designed API calls that retrieve question-relevant columns from tables.

\noindent\textbf{Reasoning with knowledge}~~~
By selecting knowledge, reasoning tasks can be solved in a more grounded and interpretable way. 
To generate an entailment tree explanation for a given hypothesis, \citet{neves-ribeiro-etal-2022-entailment} retrieves from textual premises iteratively and combines them with generation. \citet{yang-etal-2022-logicsolver} proposes a math reasoner that first retrieves highly-correlated algebraic knowledge and then passes them as prompts to improve the semantic representations for the generation task. With the recent advances in LLMs, \citet{he2022rethinking, li2023chain} retrieve from KG and KB, such as Wikidata, based on reasoning steps obtained from the chain-of-thought (CoT) prompting \citep{wei2022chain}.

\noindent\textbf{Knowledge-grounded dialogue}~~~
Dialogue generation based on relevant tables and knowledge bases has been a practical research application \citep{wu-etal-2020-diverse, li2022opera, nakamura-etal-2022-hybridialogue, gao-etal-2022-comfact, lu-etal-2023-statcan}. To tackle the challenge, \citet{li-etal-2022-knowledge} and \citet{galetzka-etal-2021-space} retrieve relevant knowledge, process it into a dense representation and incorporate it into dialogue generation. On top of dense representations, \citet{gu-etal-2020-filtering} and \citet{jung-etal-2020-attnio} leverage attention mechanisms to flexibly adjust which knowledge to depend on during generation. 
Some methods~\citep{zhang-etal-2021-kers-knowledge, dziri-etal-2021-neural, chen-etal-2020-airconcierge} first generate subgoals or responses and then use them to retrieve relevant knowledge. The retrieved knowledge then helps amend previous responses. 
Besides knowledge, \citet{cai-etal-2019-retrieval} and \citet{wu-etal-2020-improving-knowledge} improve dialogue response generation by retrieving templates or prototype dialogues {to augment inputs}. Recently, \citet{kang2023knowledge} retrieves relevant subgraphs from KGs, and then utilizes contrastive learning to ensure that the generated texts have high similarity to the subgraphs.


By retrieving from relevant sources, RAG not only improves factuality but also provides the grounding contexts while generating, thus addressing interpretability and robustness concerns. With the potential to handle more information types with recent advances in LLMs \citep{gpt4}, RAG with structured knowledge could be further enhanced. There are still challenges to be addressed. For example, there could be new designs for better retrieval systems that could promote efficient interactions suitable for diverse knowledge bases. Synthesizing this information correctly is also an open challenge, where it is hard to decide which parts need augmenting in the textual outputs.

%% file: sections/3_audio.tex
Audio RAG is useful in incorporating audio information in specific audio-language tasks, such as music captioning, music and text generation, and speech recognition. Moreover, using audio RAG for audio data augmentation has also been proven useful in mitigating the lack of audio-text training data. It could be a promising future direction \citep{li2022survey}.

\noindent\textbf{Text-audio data augmentation}~~~
For text-audio tasks, one of the most important challenges is the lack of training data on audio-text pairs. Therefore, retrieving audio and textual cues can alleviate the data scarcity problem and improve performance. In audio captioning, which aims at translating the input audio into its description, \citet{koizumi2020audio} retrieves guidance captions similar to the input audio from the training set. Then, the retrieved guidance captions are fed into a PLM to help generate new captions, which improves generation performance. To augment scarce speech translation (ST) data, \citet{zhao-etal-2023-generating} proposes SpokenVocab, a technique to convert machine translation (MT) data to synthetic ST data. To form synthetic speech, SpokenVocab retrieves and stitches audio snippets, corresponding to words in an MT sentence. Experiments show that stitched audio snippets can improve translation quality. \citet{kim2023prefix} leverages a PLM to tackle the data scarcity issue. It retrieves features from the input audio, maps them to continuous vectors using mapping networks, and uses vectors as prefixes for prefix tuning the PLM. With the additional information from retrieved audio, it outperforms previous methods. In text-to-audio generation, \citet{huang2023make} applies audio-text retrieval to get pseudo text prompts, which enhance audio generation in data-scarce scenarios. To augment the argumentation mining (AM) task in political debates, \citet{mestre-etal-2023-augmenting} integrates audio features into PLMs, which improves performance when data is scarce.

\noindent\textbf{Music captioning}~~~
Music captioning is the task of generating a text description or lyrics given the music audio. And RAG is explored to learn better audio-lyric alignment. \citet{manco2021muscaps} proposes the first music audio captioning model, MusCaps. Firstly, a pretrained multimodal encoder obtains audio representations that retrieve musical features in the input. As the pretraining bridges the gap between the audio modality and textual understanding, the method improves task performance. \citet{he2022recap} learns an audio-lyric alignment through contrastive learning, which results in a higher-quality generation of captions for music.

\noindent\textbf{Music generation}~~~
\citet{royal2020deep} uses deep neural hashing to retrieve music building blocks and then performs generation by using the current music segment to retrieve the next. In automatic speech recognition (ASR), \citet{chan2023using} uses a k-nearest neighbor (KNN) approach to retrieve external knowledge related to the audio and text embeddings. The retrieved knowledge significantly reduces domain adaptation time for ASR.


The audio modality is closely intertwined with other modalities, such as video. Therefore, recent advancements in using
audio features for text-video retrieval \citep{falcon2022feature, mithun2018learning} can benefit RAG tasks involving other modalities. Moreover, although audio-text retrieval has been a long-standing task \citep{liu2015combining, milde-etal-2016-ambient, milde-etal-2016-demonstrating}, exploring recently discovered techniques \citep{hu2022audio, lou2022audio, koepke2022audio} could lead to further improvements.

%% file: sections/3_video.tex

Retrieving video snippets for generation is used primarily in two tasks: video-grounded dialogue and video captioning. Recently, augmenting LLMs with video retrieval also demonstrates good performances, especially in few-shot settings.

\noindent\textbf{Video-grounded dialogue}~~~
Given video contexts, the model learns to engage in a relevant dialogue. \citet{pasunuru2018game} introduces a video-context, many-speaker dialogue dataset, which challenges researchers to develop visually-grounded dialogue models that generate relevant responses from live videos. Similarly, \citet{lei-etal-2020-tvqa} proposes TVQA+, a dataset that requires retrieving relevant video moments to answer textual questions about videos. Then, it proposes a unified framework that encodes video segments into representations, uses an attention mechanism to locate relevant information, and produces textual answers. To better perform visually-grounded dialogue tasks, \citet{le-etal-2020-bist} retrieves visual cues from prior user queries. The cues are then used as contextual information to construct relevant responses. On video QA, it substantially outperforms prior approaches. Recently, \citet{le2022vgnmn} extracts visual cues from the video to augment video-grounded dialogues. The video retrieval is performed with neural module networks, which are instantiated with entities and actions in previous dialogues. 

\noindent\textbf{Video captioning}~~~
{Sharing a similar motivation to RAG, }\citet{long2018video} first proposes to use attention layers to automatically select the most salient visual or semantic features and use them to augment caption generation. As a result, it stably outperforms previous methods. \citep{whitehead-etal-2018-incorporating} then develops a retrieval-based approach for video description generation. For news videos, it retrieves topically related news documents and then generates a description using a knowledge-aware video description network.

\noindent\textbf{LLM augmentation}~~
\citet{wang2022language} attempts to augment an LLM to generalize to various video-to-text tasks from a few examples. As the LLMs cannot accept video inputs, it first translates video contents into attributes using image-language models and then prompts the retrieved content to instruct the LLM. It demonstrates good few-shot performances on a wide range of video-language tasks.

Currently, the video-text research bottleneck mainly lies in the representation gap between different modalities. Research has been attempting to learn a better mapping between video-text via joint learning \citep{xu2015jointly, sun2019videobert}. Recent studies on dense video representation learning can also be useful for future video RAG. Besides, some papers~\citep{yang23vid2seq,wang2021t2vlad} try to introduce fine-grained interaction between different modalities {to learn better aligned representations}. \citet{zeng2022socratic} encourages multiple pretrained models in different modalities to exchange information with each other in a zero-shot manner. Most recently, \citet{zhang2023video} trains Video-Llama to better align pretrained video and audio encoders with LLM's embedding space.

%% file: sections/4_future.tex
With the development of multi-modal LLMs, retrieving multimodal information to augment text generation will be a promising direction to better ground textual generation in real-world contexts, contributing towards building a model that is fully aware and can better interact with the world. Specifically, we describe some potential directions that can be of benefit to the community.

\subsection{Retrieval Augmented Multimodal Reasoning}
\vspace{-1ex}


One potential application of multimodal RAG
is multimodal reasoning. \citet{lu2022learn} first introduces ScienceQA, a large-scale multimodal science question dataset annotated with lectures and explanations. Then, \citet{zhang2023multimodal} proposes Multimodal Chain-of-Thought (Multimodal-CoT) which incorporates language and vision modalities into a two-stage (rationale generation and answer inference) framework, surpassing GPT-3.5 by a large margin with a much smaller fine-tuned model. Similar to \citet{zhang2023multimodal}, kosmos-1 \citep{huang2023language} breaks down multimodal reasoning into two steps. It first generates intermediate content as the rationale based on visual information and then uses the generated rationale to induce the result. However, both methods may have difficulties understanding certain types of images (e.g., maps), which could be mitigated by retrieving informative image-text pairs. 

\vspace{-1ex}
\subsection{Building a Multimodal Knowledge Index}

In order to facilitate multimodal RAG, one of the most fundamental aspects is building a multimodal knowledge index. The goal is twofold: Firstly, dense representations should support low storage, dynamic updating of the knowledge base, and accurate search. Secondly, it could enable faster search speed with the help of local sensitive hashing~\citep{data-mining}, which combats scaling and robustness concerns when the knowledge base is scaled up extremely.

Currently, the dense representations for text snippets are widely studied for documents~\cite{karpukhin-etal-2020-dense,gao-callan-2021-condenser,gao-etal-2021-simcse}, entities~\citep{sciavolino-etal-2021-simple,lee-etal-2021-learning-dense}, and images~\cite{clip}. Besides, there are studies optimizing dense representations in an end-to-end manner~\cite{lewis2020retrieval}.
Nevertheless, few papers~\citep{chen2022murag} have explored building 
a multimodal index at the same time for downstream generation tasks.
How to map a multimodal knowledge index into a unified space remains a long-term challenge.

\vspace{-1ex}
\subsection{Pretraining with Multimodal Retrieval}

To better align the abilities to handle different modalities in a pre-trained model, future work could be built on employing retrieval-based approaches during pre-training. Currently, some methods fine-tune the pre-trained generative model to learn to retrieve from different modalities.
For example, LaMDA~\citep{thoppilan2022lamda} calls an external toolset for fine-tuning, including an information retrieval system.
Similarly, during fine-tuning, Toolformer \citep{schick2023toolformer} augments models with API calls to tools including a QA system and a Wikipedia search engine. 

When similar retrieval abilities are leveraged during pretraining, the generative models can interact with retrieval tools much better. Then, instead of relying solely on internal weights, they could effectively use an external base to output more grounded information, provide relevant contexts to users, and update their information accordingly. Such pretraining techniques would also greatly improve robustness for out-of-domain tasks. As an example, \citet{guu2020realm} augments pretraining with an external knowledge retriever, which outperforms previous methods. \citet{aiello2023jointly} employs multimodal retrieval augmentation while training, resulting in a first-of-its-kind large multimodal model that can coherently generate long-form content with interleaved texts and images.

To incorporate retrieval with pretraining, there remains the challenge of developing appropriate datasets labeled with retrieval API calls. To tackle this challenge, LaMDA \citep{thoppilan2022lamda} uses labels developed by human annotators, which could be expensive to collect. Toolformer \citep{schick2023toolformer} uses a sampling and filtering approach for automatic labeling, which is inexpensive but could induce bias. A potential solution is to use a neuro-symbolic approach~\citep{davoudi2021toward}, which uses prototype learning and deep-KNN to find nearest neighbors during training.

%% file: sections/5_conclusion.tex
This survey reviews research that augments generative models by retrieving multi-modal information. Specifically, we categorize the current domain into enhancing with different modalities, including image, code, structured knowledge, speech, and video. 
With the emergence of large multi-modal models, we believe that this survey could serve as a comprehensive overview of an emerging and promising field. Moreover, we hope it could encourage future research in the domain, including retrieval-augmented multimodal reasoning, building a multi-modal knowledge index, and combining retrieval with pretraining.

%% file: sections/app.tex
\subsection{Search Criteria and Results}
\label{subsec:search}

For searching the ACL anthology articles, we use a keyword search over titles and abstracts. We strictly enforce the keyword ``retriev''. Then, we enforce either ``generat'' or ``ground'' to appear. For each modality, we then add modality-specific keywords: ``image'' for the image modality, ``code'' for the code modality, any one from ``structured knowledge/table/database/knowledge graph'' for the structured knowledge modality, any one from ``audio/speech'' for the audio modality, and ``video'' for the video modality.

For searching on Google Scholar, we add the keyword ``language models'' to select more NLP-related articles. We then perform manual filtering on the top 3 pages of returned results.

\input{sections/paper_stat_table}
\begin{figure}[t!]
    \centering
    \includegraphics[width=0.5\textwidth]{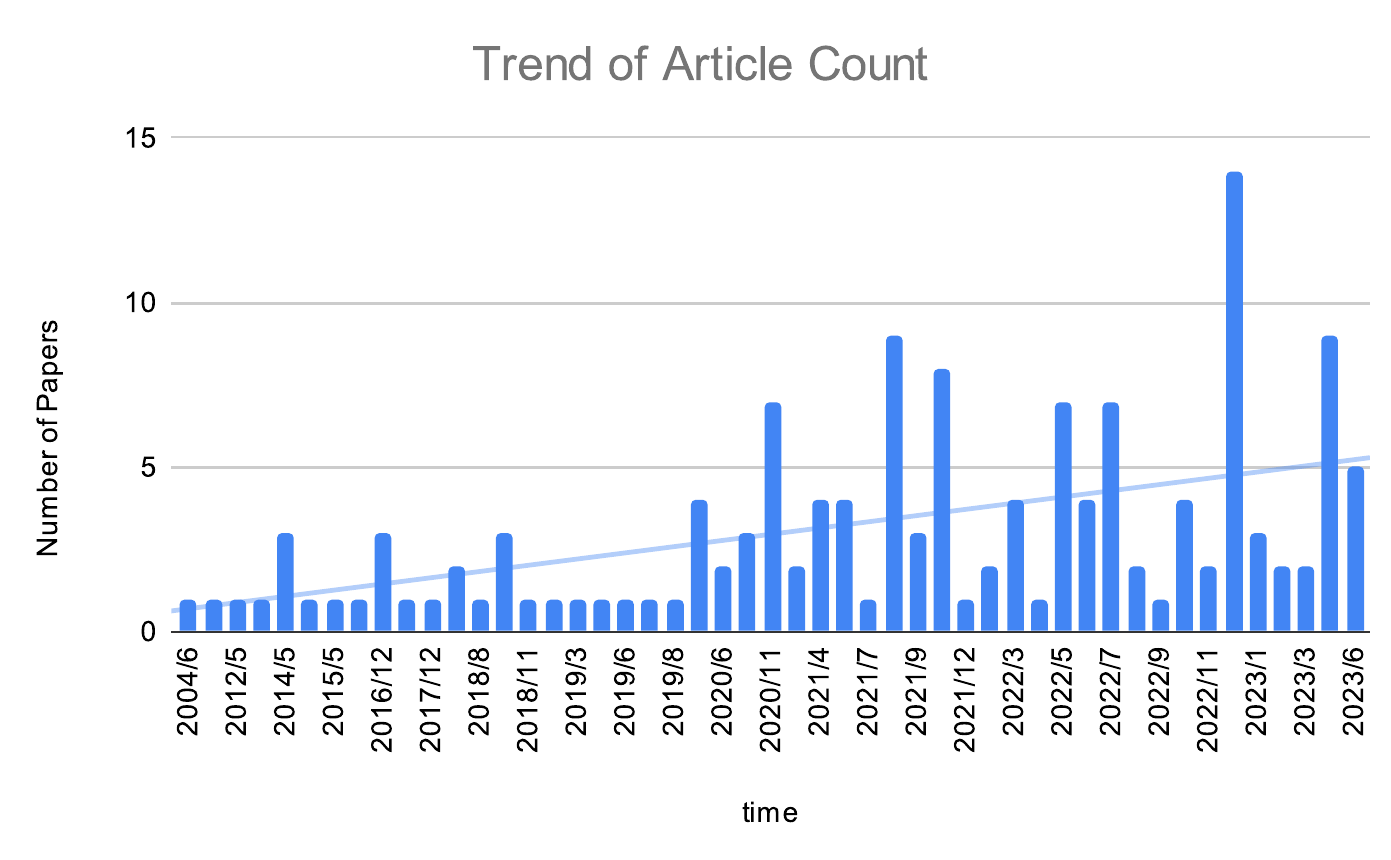}
  \caption{Paper trend analysis}
  \label{fig:trend_fig}
\end{figure}
The number of retrieved and analyzed research papers can be found in \Cref{tab:paper_stats}.

A trend analysis of how the number of papers change across time is shown in \Cref{fig:trend_fig} We could observe that the domain of multimodal retrieval-augmented generation has indeed developed a lot recently, with peaks reached around end of 2022. The observation is consistent with our hypothesis that multimodal RAG is especially important and helpful in the age of large-scale general-purpose models.

%% file: sections/paper_stat_table.tex
\begin{table}[t!]
\centering
\begin{adjustbox}{width=\columnwidth,center}
\begin{tabular}{ccccc}
\toprule
\textbf{Modality} & \textbf{ACL} & \textbf{Google} & \textbf{Total analyzed}\\
\midrule
Image & (67) 17 & 6 & 23 \\
Code & (177) 9 & 24 & 33 \\
Structured & (108) 44 & 11 & 55 \\
Audio & (17) 6 & 14 & 20 \\
Video & (22) 7 & 7 & 14 \\
\midrule
Total & (291) 83 & 62 & 145 \\
\bottomrule
\end{tabular}
\end{adjustbox}
\caption{Paper statistics. Number in parenthesis is the number before manual filtering. ``Google'' represents searching on google scholar and manually filtering. ``Total analyzed'' represents the number of total papers after manual filtering}
\label{tab:paper_stats}
\end{table}

%% file: emnlp2023.bbl
\begin{thebibliography}{205}
\expandafter\ifx\csname natexlab\endcsname\relax\def\natexlab#1{#1}\fi

\bibitem[{Acosta et~al.(2022)Acosta, Falcone, Rajpurkar, and
  Topol}]{acosta2022multimodal}
Juli{\'a}n~N Acosta, Guido~J Falcone, Pranav Rajpurkar, and Eric~J Topol. 2022.
\newblock Multimodal biomedical ai.
\newblock \emph{Nature Medicine}, 28(9):1773--1784.

\bibitem[{Aggarwal et~al.(2021)Aggarwal, Mandowara, Agrawal, Khandelwal,
  Singla, and Garg}]{aggarwal-etal-2021-explanations}
Shourya Aggarwal, Divyanshu Mandowara, Vishwajeet Agrawal, Dinesh Khandelwal,
  Parag Singla, and Dinesh Garg. 2021.
\newblock \href {https://doi.org/10.18653/v1/2021.acl-long.238} {{E}xplanations
  for {C}ommonsense{QA}: {N}ew {D}ataset and {M}odels}.
\newblock In \emph{Proceedings of the 59th Annual Meeting of the Association
  for Computational Linguistics and the 11th International Joint Conference on
  Natural Language Processing (Volume 1: Long Papers)}, pages 3050--3065,
  Online. Association for Computational Linguistics.

\bibitem[{Aghajanyan et~al.(2022)Aghajanyan, Huang, Ross, Karpukhin, Xu, Goyal,
  Okhonko, Joshi, Ghosh, Lewis, and Zettlemoyer}]{CM3}
Armen Aghajanyan, Bernie Huang, Candace Ross, Vladimir Karpukhin, Hu~Xu, Naman
  Goyal, Dmytro Okhonko, Mandar Joshi, Gargi Ghosh, Mike Lewis, and Luke
  Zettlemoyer. 2022.
\newblock {CM3:} {A} causal masked multimodal model of the internet.
\newblock \emph{CoRR}, abs/2201.07520.

\bibitem[{Aiello et~al.(2023)Aiello, Yu, Nie, Aghajanyan, and
  Oguz}]{aiello2023jointly}
Emanuele Aiello, Lili Yu, Yixin Nie, Armen Aghajanyan, and Barlas Oguz. 2023.
\newblock Jointly training large autoregressive multimodal models.
\newblock \emph{arXiv preprint arXiv:2309.15564}.

\bibitem[{Aksitov et~al.(2023)Aksitov, Chang, Reitter, Shakeri, and
  Sung}]{aksitov2023characterizing}
Renat Aksitov, Chung-Ching Chang, David Reitter, Siamak Shakeri, and Yunhsuan
  Sung. 2023.
\newblock \href {http://arxiv.org/abs/2302.05578} {Characterizing attribution
  and fluency tradeoffs for retrieval-augmented large language models}.

\bibitem[{Alayrac et~al.(2022{\natexlab{a}})Alayrac, Donahue, Luc, Miech, Barr,
  Hasson, Lenc, Mensch, Millican, Reynolds et~al.}]{alayracflamingo}
Jean-Baptiste Alayrac, Jeff Donahue, Pauline Luc, Antoine Miech, Iain Barr,
  Yana Hasson, Karel Lenc, Arthur Mensch, Katherine Millican, Malcolm Reynolds,
  et~al. 2022{\natexlab{a}}.
\newblock Flamingo: a visual language model for few-shot learning.
\newblock In \emph{Advances in Neural Information Processing Systems}.

\bibitem[{Alayrac et~al.(2022{\natexlab{b}})Alayrac, Donahue, Luc, Miech, Barr,
  Hasson, Lenc, Mensch, Millican, Reynolds, Ring, Rutherford, Cabi, Han, Gong,
  Samangooei, Monteiro, Menick, Borgeaud, Brock, Nematzadeh, Sharifzadeh,
  Binkowski, Barreira, Vinyals, Zisserman, and Simonyan}]{flamingo}
Jean{-}Baptiste Alayrac, Jeff Donahue, Pauline Luc, Antoine Miech, Iain Barr,
  Yana Hasson, Karel Lenc, Arthur Mensch, Katie Millican, Malcolm Reynolds,
  Roman Ring, Eliza Rutherford, Serkan Cabi, Tengda Han, Zhitao Gong, Sina
  Samangooei, Marianne Monteiro, Jacob Menick, Sebastian Borgeaud, Andrew
  Brock, Aida Nematzadeh, Sahand Sharifzadeh, Mikolaj Binkowski, Ricardo
  Barreira, Oriol Vinyals, Andrew Zisserman, and Karen Simonyan.
  2022{\natexlab{b}}.
\newblock Flamingo: a visual language model for few-shot learning.
\newblock \emph{CoRR}, abs/2204.14198.

\bibitem[{Baek et~al.(2023)Baek, Aji, and Saffari}]{baek2023knowledge}
Jinheon Baek, Alham~Fikri Aji, and Amir Saffari. 2023.
\newblock Knowledge-augmented language model prompting for zero-shot knowledge
  graph question answering.
\newblock \emph{arXiv preprint arXiv:2306.04136}.

\bibitem[{Baltru{\v{s}}aitis et~al.(2018)Baltru{\v{s}}aitis, Ahuja, and
  Morency}]{baltruvsaitis2018multimodal}
Tadas Baltru{\v{s}}aitis, Chaitanya Ahuja, and Louis-Philippe Morency. 2018.
\newblock Multimodal machine learning: A survey and taxonomy.
\newblock \emph{IEEE transactions on pattern analysis and machine
  intelligence}, 41(2):423--443.

\bibitem[{Borgeaud et~al.(2022)Borgeaud, Mensch, Hoffmann, Cai, Rutherford,
  Millican, Van Den~Driessche, Lespiau, Damoc, Clark
  et~al.}]{borgeaud2022improving}
Sebastian Borgeaud, Arthur Mensch, Jordan Hoffmann, Trevor Cai, Eliza
  Rutherford, Katie Millican, George~Bm Van Den~Driessche, Jean-Baptiste
  Lespiau, Bogdan Damoc, Aidan Clark, et~al. 2022.
\newblock Improving language models by retrieving from trillions of tokens.
\newblock In \emph{International conference on machine learning}, pages
  2206--2240. PMLR.

\bibitem[{Brown et~al.(2020)Brown, Mann, Ryder, Subbiah, Kaplan, Dhariwal,
  Neelakantan, Shyam, Sastry, Askell et~al.}]{brown2020language}
Tom Brown, Benjamin Mann, Nick Ryder, Melanie Subbiah, Jared~D Kaplan, Prafulla
  Dhariwal, Arvind Neelakantan, Pranav Shyam, Girish Sastry, Amanda Askell,
  et~al. 2020.
\newblock Language models are few-shot learners.
\newblock \emph{Advances in neural information processing systems},
  33:1877--1901.

\bibitem[{Cai et~al.(2019{\natexlab{a}})Cai, Wang, Bi, Tu, Liu, Lam, and
  Shi}]{cai-etal-2019-skeleton}
Deng Cai, Yan Wang, Wei Bi, Zhaopeng Tu, Xiaojiang Liu, Wai Lam, and Shuming
  Shi. 2019{\natexlab{a}}.
\newblock \href {https://doi.org/10.18653/v1/N19-1124} {Skeleton-to-response:
  Dialogue generation guided by retrieval memory}.
\newblock In \emph{Proceedings of the 2019 Conference of the North {A}merican
  Chapter of the Association for Computational Linguistics: Human Language
  Technologies, Volume 1 (Long and Short Papers)}, pages 1219--1228,
  Minneapolis, Minnesota. Association for Computational Linguistics.

\bibitem[{Cai et~al.(2019{\natexlab{b}})Cai, Wang, Bi, Tu, Liu, and
  Shi}]{cai-etal-2019-retrieval}
Deng Cai, Yan Wang, Wei Bi, Zhaopeng Tu, Xiaojiang Liu, and Shuming Shi.
  2019{\natexlab{b}}.
\newblock \href {https://doi.org/10.18653/v1/D19-1195} {Retrieval-guided
  dialogue response generation via a matching-to-generation framework}.
\newblock In \emph{Proceedings of the 2019 Conference on Empirical Methods in
  Natural Language Processing and the 9th International Joint Conference on
  Natural Language Processing (EMNLP-IJCNLP)}, pages 1866--1875, Hong Kong,
  China. Association for Computational Linguistics.

\bibitem[{Chan et~al.(2023)Chan, Ghosh, Rastrow, and
  Hoffmeister}]{chan2023using}
David~M Chan, Shalini Ghosh, Ariya Rastrow, and Bj{\"o}rn Hoffmeister. 2023.
\newblock Using external off-policy speech-to-text mappings in contextual
  end-to-end automated speech recognition.
\newblock \emph{arXiv preprint arXiv:2301.02736}.

\bibitem[{Chen et~al.(2020)Chen, Wang, Chang, Juan, Wei, and
  Pan}]{chen-etal-2020-airconcierge}
Chieh-Yang Chen, Pei-Hsin Wang, Shih-Chieh Chang, Da-Cheng Juan, Wei Wei, and
  Jia-Yu Pan. 2020.
\newblock \href {https://doi.org/10.18653/v1/2020.findings-emnlp.79}
  {{A}ir{C}oncierge: Generating task-oriented dialogue via efficient
  large-scale knowledge retrieval}.
\newblock In \emph{Findings of the Association for Computational Linguistics:
  EMNLP 2020}, pages 884--897, Online. Association for Computational
  Linguistics.

\bibitem[{Chen et~al.(2021)Chen, Tworek, Jun, Yuan, Pinto, Kaplan, Edwards,
  Burda, Joseph, Brockman et~al.}]{codex}
Mark Chen, Jerry Tworek, Heewoo Jun, Qiming Yuan, Henrique Ponde de~Oliveira
  Pinto, Jared Kaplan, Harri Edwards, Yuri Burda, Nicholas Joseph, Greg
  Brockman, et~al. 2021.
\newblock Evaluating large language models trained on code.
\newblock \emph{arXiv preprint arXiv:2107.03374}.

\bibitem[{Chen et~al.(2022{\natexlab{a}})Chen, Hu, Chen, Verga, and
  Cohen}]{chen2022murag}
Wenhu Chen, Hexiang Hu, Xi~Chen, Pat Verga, and William Cohen.
  2022{\natexlab{a}}.
\newblock \href {https://aclanthology.org/2022.emnlp-main.375} {{M}u{RAG}:
  Multimodal retrieval-augmented generator for open question answering over
  images and text}.
\newblock In \emph{EMNLP}, pages 5558--5570. ACL.

\bibitem[{Chen et~al.(2022{\natexlab{b}})Chen, Hu, Chen, Verga, and
  Cohen}]{chen-etal-2022-murag}
Wenhu Chen, Hexiang Hu, Xi~Chen, Pat Verga, and William Cohen.
  2022{\natexlab{b}}.
\newblock \href {https://aclanthology.org/2022.emnlp-main.375} {{M}u{RAG}:
  Multimodal retrieval-augmented generator for open question answering over
  images and text}.
\newblock In \emph{Proceedings of the 2022 Conference on Empirical Methods in
  Natural Language Processing}, pages 5558--5570, Abu Dhabi, United Arab
  Emirates. Association for Computational Linguistics.

\bibitem[{Chen et~al.(2022{\natexlab{c}})Chen, Hu, Saharia, and
  Cohen}]{chen2022re}
Wenhu Chen, Hexiang Hu, Chitwan Saharia, and William~W Cohen.
  2022{\natexlab{c}}.
\newblock Re-imagen: Retrieval-augmented text-to-image generator.
\newblock \emph{arXiv preprint arXiv:2209.14491}.

\bibitem[{Chen et~al.(2022{\natexlab{d}})Chen, Ma, Wang, and
  Cohen}]{chen2022program}
Wenhu Chen, Xueguang Ma, Xinyi Wang, and William~W Cohen. 2022{\natexlab{d}}.
\newblock Program of thoughts prompting: Disentangling computation from
  reasoning for numerical reasoning tasks.
\newblock \emph{arXiv preprint arXiv:2211.12588}.

\bibitem[{Cheng et~al.(2020)Cheng, Wu, Bing, Zhang, Jie, Lu, and
  Si}]{cheng-etal-2020-ent}
Liying Cheng, Dekun Wu, Lidong Bing, Yan Zhang, Zhanming Jie, Wei Lu, and Luo
  Si. 2020.
\newblock \href {https://doi.org/10.18653/v1/2020.emnlp-main.90} {{ENT}-{DESC}:
  Entity description generation by exploring knowledge graph}.
\newblock In \emph{Proceedings of the 2020 Conference on Empirical Methods in
  Natural Language Processing (EMNLP)}, pages 1187--1197, Online. Association
  for Computational Linguistics.

\bibitem[{Cheng et~al.(2023)Cheng, Xie, Shi, Li, Nadkarni, Hu, Xiong, Radev,
  Ostendorf, Zettlemoyer, Smith, and Yu}]{cheng2022binding}
Zhoujun Cheng, Tianbao Xie, Peng Shi, Chengzu Li, Rahul Nadkarni, Yushi Hu,
  Caiming Xiong, Dragomir Radev, Mari Ostendorf, Luke Zettlemoyer, Noah~A.
  Smith, and Tao Yu. 2023.
\newblock Binding language models in symbolic languages.
\newblock \emph{ICLR}.

\bibitem[{Cho et~al.(2022)Cho, Yoon, Kale, Dernoncourt, Bui, and
  Bansal}]{cho-etal-2022-fine}
Jaemin Cho, Seunghyun Yoon, Ajinkya Kale, Franck Dernoncourt, Trung Bui, and
  Mohit Bansal. 2022.
\newblock \href {https://doi.org/10.18653/v1/2022.findings-naacl.39}
  {Fine-grained image captioning with {CLIP} reward}.
\newblock In \emph{Findings of the Association for Computational Linguistics:
  NAACL 2022}, pages 517--527, Seattle, United States. Association for
  Computational Linguistics.

\bibitem[{Chowdhery et~al.(2022)Chowdhery, Narang, Devlin, Bosma, Mishra,
  Roberts, Barham, Chung, Sutton, Gehrmann et~al.}]{chowdhery2022palm}
Aakanksha Chowdhery, Sharan Narang, Jacob Devlin, Maarten Bosma, Gaurav Mishra,
  Adam Roberts, Paul Barham, Hyung~Won Chung, Charles Sutton, Sebastian
  Gehrmann, et~al. 2022.
\newblock Palm: Scaling language modeling with pathways.
\newblock \emph{arXiv preprint arXiv:2204.02311}.

\bibitem[{Clement et~al.(2021)Clement, Lu, Liu, Tufano, Drain, Duan,
  Sundaresan, and Svyatkovskiy}]{clement2021long}
Colin~B Clement, Shuai Lu, Xiaoyu Liu, Michele Tufano, Dawn Drain, Nan Duan,
  Neel Sundaresan, and Alexey Svyatkovskiy. 2021.
\newblock Long-range modeling of source code files with ewash: Extended window
  access by syntax hierarchy.
\newblock \emph{arXiv preprint arXiv:2109.08780}.

\bibitem[{Crowson et~al.(2022)Crowson, Biderman, Kornis, Stander, Hallahan,
  Castricato, and Raff}]{crowson2022vqgan}
Katherine Crowson, Stella Biderman, Daniel Kornis, Dashiell Stander, Eric
  Hallahan, Louis Castricato, and Edward Raff. 2022.
\newblock Vqgan-clip: Open domain image generation and editing with natural
  language guidance.
\newblock In \emph{Computer Vision--ECCV 2022: 17th European Conference, Tel
  Aviv, Israel, October 23--27, 2022, Proceedings, Part XXXVII}, pages 88--105.
  Springer.

\bibitem[{Davoudi and Komeili(2021)}]{davoudi2021toward}
Seyed~Omid Davoudi and Majid Komeili. 2021.
\newblock Toward faithful case-based reasoning through learning prototypes in a
  nearest neighbor-friendly space.
\newblock In \emph{International Conference on Learning Representations}.

\bibitem[{Ding et~al.(2022)Ding, Wang, Ahmad, Ramanathan, Nallapati, Bhatia,
  Roth, and Xiang}]{ding2022cocomic}
Yangruibo Ding, Zijian Wang, Wasi~Uddin Ahmad, Murali~Krishna Ramanathan,
  Ramesh Nallapati, Parminder Bhatia, Dan Roth, and Bing Xiang. 2022.
\newblock Cocomic: Code completion by jointly modeling in-file and cross-file
  context.
\newblock \emph{arXiv preprint arXiv:2212.10007}.

\bibitem[{Dou et~al.(2022)Dou, Kamath, Gan, Zhang, Wang, Li, Liu, Liu, LeCun,
  Peng et~al.}]{dou2022coarse}
Zi-Yi Dou, Aishwarya Kamath, Zhe Gan, Pengchuan Zhang, Jianfeng Wang, Linjie
  Li, Zicheng Liu, Ce~Liu, Yann LeCun, Nanyun Peng, et~al. 2022.
\newblock Coarse-to-fine vision-language pre-training with fusion in the
  backbone.
\newblock \emph{arXiv preprint arXiv:2206.07643}.

\bibitem[{Driess et~al.(2023)Driess, Xia, Sajjadi, Lynch, Chowdhery, Ichter,
  Wahid, Tompson, Vuong, Yu et~al.}]{driess2023palm}
Danny Driess, Fei Xia, Mehdi~SM Sajjadi, Corey Lynch, Aakanksha Chowdhery,
  Brian Ichter, Ayzaan Wahid, Jonathan Tompson, Quan Vuong, Tianhe Yu, et~al.
  2023.
\newblock Palm-e: An embodied multimodal language model.
\newblock \emph{arXiv preprint arXiv:2303.03378}.

\bibitem[{Dziri et~al.(2021)Dziri, Madotto, Za{\"\i}ane, and
  Bose}]{dziri-etal-2021-neural}
Nouha Dziri, Andrea Madotto, Osmar Za{\"\i}ane, and Avishek~Joey Bose. 2021.
\newblock \href {https://doi.org/10.18653/v1/2021.emnlp-main.168} {Neural path
  hunter: Reducing hallucination in dialogue systems via path grounding}.
\newblock In \emph{Proceedings of the 2021 Conference on Empirical Methods in
  Natural Language Processing}, pages 2197--2214, Online and Punta Cana,
  Dominican Republic. Association for Computational Linguistics.

\bibitem[{Falcon et~al.(2022)Falcon, Serra, and Lanz}]{falcon2022feature}
Alex Falcon, Giuseppe Serra, and Oswald Lanz. 2022.
\newblock A feature-space multimodal data augmentation technique for text-video
  retrieval.
\newblock In \emph{Proceedings of the 30th ACM International Conference on
  Multimedia}, pages 4385--4394.

\bibitem[{Fan et~al.(2019)Fan, Gardent, Braud, and Bordes}]{fan2019using}
Angela Fan, Claire Gardent, Chlo{\'e} Braud, and Antoine Bordes. 2019.
\newblock Using local knowledge graph construction to scale seq2seq models to
  multi-document inputs.
\newblock \emph{arXiv preprint arXiv:1910.08435}.

\bibitem[{Fan et~al.(2021)Fan, Gardent, Braud, and
  Bordes}]{fan-etal-2021-augmenting}
Angela Fan, Claire Gardent, Chlo{\'e} Braud, and Antoine Bordes. 2021.
\newblock \href {https://doi.org/10.1162/tacl_a_00356} {Augmenting transformers
  with {KNN}-based composite memory for dialog}.
\newblock \emph{Transactions of the Association for Computational Linguistics},
  9:82--99.

\bibitem[{Fang and Feng(2022)}]{fang-feng-2022-neural}
Qingkai Fang and Yang Feng. 2022.
\newblock \href {https://doi.org/10.18653/v1/2022.acl-long.390} {Neural machine
  translation with phrase-level universal visual representations}.
\newblock In \emph{Proceedings of the 60th Annual Meeting of the Association
  for Computational Linguistics (Volume 1: Long Papers)}, pages 5687--5698,
  Dublin, Ireland. Association for Computational Linguistics.

\bibitem[{Frisoni et~al.(2022)Frisoni, Mizutani, Moro, and
  Valgimigli}]{frisoni-etal-2022-bioreader}
Giacomo Frisoni, Miki Mizutani, Gianluca Moro, and Lorenzo Valgimigli. 2022.
\newblock \href {https://aclanthology.org/2022.emnlp-main.390} {{B}io{R}eader:
  a retrieval-enhanced text-to-text transformer for biomedical literature}.
\newblock In \emph{Proceedings of the 2022 Conference on Empirical Methods in
  Natural Language Processing}, pages 5770--5793, Abu Dhabi, United Arab
  Emirates. Association for Computational Linguistics.

\bibitem[{Galetzka et~al.(2021)Galetzka, Rose, Schlangen, and
  Lehmann}]{galetzka-etal-2021-space}
Fabian Galetzka, Jewgeni Rose, David Schlangen, and Jens Lehmann. 2021.
\newblock \href {https://doi.org/10.18653/v1/2021.acl-long.546} {Space
  efficient context encoding for non-task-oriented dialogue generation with
  graph attention transformer}.
\newblock In \emph{Proceedings of the 59th Annual Meeting of the Association
  for Computational Linguistics and the 11th International Joint Conference on
  Natural Language Processing (Volume 1: Long Papers)}, pages 7028--7041,
  Online. Association for Computational Linguistics.

\bibitem[{Gao et~al.(2020)Gao, Li, Chen, and Zhang}]{gao2020survey}
Jing Gao, Peng Li, Zhikui Chen, and Jianing Zhang. 2020.
\newblock A survey on deep learning for multimodal data fusion.
\newblock \emph{Neural Computation}, 32(5):829--864.

\bibitem[{Gao and Callan(2021)}]{gao-callan-2021-condenser}
Luyu Gao and Jamie Callan. 2021.
\newblock \href {https://doi.org/10.18653/v1/2021.emnlp-main.75} {Condenser: a
  pre-training architecture for dense retrieval}.
\newblock In \emph{Proceedings of the 2021 Conference on Empirical Methods in
  Natural Language Processing}, pages 981--993, Online and Punta Cana,
  Dominican Republic. Association for Computational Linguistics.

\bibitem[{Gao et~al.(2022{\natexlab{a}})Gao, Madaan, Zhou, Alon, Liu, Yang,
  Callan, and Neubig}]{gao2022pal}
Luyu Gao, Aman Madaan, Shuyan Zhou, Uri Alon, Pengfei Liu, Yiming Yang, Jamie
  Callan, and Graham Neubig. 2022{\natexlab{a}}.
\newblock Pal: Program-aided language models.
\newblock \emph{arXiv preprint arXiv:2211.10435}.

\bibitem[{Gao et~al.(2022{\natexlab{b}})Gao, Hwang, Kanno, Wakaki, Mitsufuji,
  and Bosselut}]{gao-etal-2022-comfact}
Silin Gao, Jena~D. Hwang, Saya Kanno, Hiromi Wakaki, Yuki Mitsufuji, and
  Antoine Bosselut. 2022{\natexlab{b}}.
\newblock \href {https://aclanthology.org/2022.findings-emnlp.120}
  {{C}om{F}act: A benchmark for linking contextual commonsense knowledge}.
\newblock In \emph{Findings of the Association for Computational Linguistics:
  EMNLP 2022}, pages 1656--1675, Abu Dhabi, United Arab Emirates. Association
  for Computational Linguistics.

\bibitem[{Gao et~al.(2021)Gao, Yao, and Chen}]{gao-etal-2021-simcse}
Tianyu Gao, Xingcheng Yao, and Danqi Chen. 2021.
\newblock \href {https://doi.org/10.18653/v1/2021.emnlp-main.552} {{S}im{CSE}:
  Simple contrastive learning of sentence embeddings}.
\newblock In \emph{Proceedings of the 2021 Conference on Empirical Methods in
  Natural Language Processing}, pages 6894--6910, Online and Punta Cana,
  Dominican Republic. Association for Computational Linguistics.

\bibitem[{Gu et~al.(2020)Gu, Ling, Liu, Chen, and Zhu}]{gu-etal-2020-filtering}
Jia-Chen Gu, Zhenhua Ling, Quan Liu, Zhigang Chen, and Xiaodan Zhu. 2020.
\newblock \href {https://doi.org/10.18653/v1/2020.findings-emnlp.127}
  {Filtering before iteratively referring for knowledge-grounded response
  selection in retrieval-based chatbots}.
\newblock In \emph{Findings of the Association for Computational Linguistics:
  EMNLP 2020}, pages 1412--1422, Online. Association for Computational
  Linguistics.

\bibitem[{Gu et~al.(2018)Gu, Wang, Cho, and Li}]{gu2018search}
Jiatao Gu, Yong Wang, Kyunghyun Cho, and Victor~OK Li. 2018.
\newblock Search engine guided neural machine translation.
\newblock In \emph{AAAI}, volume~32.

\bibitem[{Gu et~al.(2019)Gu, Yuqiao, and Wei}]{gu-etal-2019-extract}
Yunfan Gu, Yang Yuqiao, and Zhongyu Wei. 2019.
\newblock \href {https://doi.org/10.18653/v1/D19-5514} {Extract, transform and
  filling: A pipeline model for question paraphrasing based on template}.
\newblock In \emph{Proceedings of the 5th Workshop on Noisy User-generated Text
  (W-NUT 2019)}, pages 109--114, Hong Kong, China. Association for
  Computational Linguistics.

\bibitem[{Gui et~al.(2022)Gui, Wang, Huang, Hauptmann, Bisk, and
  Gao}]{gui-etal-2022-kat}
Liangke Gui, Borui Wang, Qiuyuan Huang, Alexander Hauptmann, Yonatan Bisk, and
  Jianfeng Gao. 2022.
\newblock \href {https://doi.org/10.18653/v1/2022.naacl-main.70} {{KAT}: A
  knowledge augmented transformer for vision-and-language}.
\newblock In \emph{Proceedings of the 2022 Conference of the North American
  Chapter of the Association for Computational Linguistics: Human Language
  Technologies}, pages 956--968, Seattle, United States. Association for
  Computational Linguistics.

\bibitem[{Guo et~al.(2022)Guo, Lu, Duan, Wang, Zhou, and Yin}]{unixcoder}
Daya Guo, Shuai Lu, Nan Duan, Yanlin Wang, Ming Zhou, and Jian Yin. 2022.
\newblock \href {https://doi.org/10.18653/v1/2022.acl-long.499} {Unixcoder:
  Unified cross-modal pre-training for code representation}.
\newblock In \emph{Proceedings of the 60th Annual Meeting of the Association
  for Computational Linguistics (Volume 1: Long Papers), {ACL} 2022, Dublin,
  Ireland, May 22-27, 2022}, pages 7212--7225. Association for Computational
  Linguistics.

\bibitem[{Gupta et~al.(2018)Gupta, Chinnakotla, and
  Shrivastava}]{gupta-etal-2018-retrieve}
Vishal Gupta, Manoj Chinnakotla, and Manish Shrivastava. 2018.
\newblock \href {https://doi.org/10.18653/v1/W18-5504} {Retrieve and re-rank: A
  simple and effective {IR} approach to simple question answering over
  knowledge graphs}.
\newblock In \emph{Proceedings of the First Workshop on Fact Extraction and
  {VER}ification ({FEVER})}, pages 22--27, Brussels, Belgium. Association for
  Computational Linguistics.

\bibitem[{Guu et~al.(2020{\natexlab{a}})Guu, Lee, Tung, Pasupat, and
  Chang}]{guu2020realm}
Kelvin Guu, Kenton Lee, Zora Tung, Panupong Pasupat, and Ming-Wei Chang.
  2020{\natexlab{a}}.
\newblock \href {http://arxiv.org/abs/2002.08909} {Realm: Retrieval-augmented
  language model pre-training}.

\bibitem[{Guu et~al.(2020{\natexlab{b}})Guu, Lee, Tung, Pasupat, and
  Chang}]{guu2020retrieval}
Kelvin Guu, Kenton Lee, Zora Tung, Panupong Pasupat, and Mingwei Chang.
  2020{\natexlab{b}}.
\newblock Retrieval augmented language model pre-training.
\newblock In \emph{International conference on machine learning}, pages
  3929--3938. PMLR.

\bibitem[{Hashimoto et~al.(2018)Hashimoto, Guu, Oren, and
  Liang}]{hashimoto2018retrieve}
Tatsunori~B Hashimoto, Kelvin Guu, Yonatan Oren, and Percy~S Liang. 2018.
\newblock A retrieve-and-edit framework for predicting structured outputs.
\newblock \emph{NeurIPS}, 31.

\bibitem[{Hayati et~al.(2018)Hayati, Olivier, Avvaru, Yin, Tomasic, and
  Neubig}]{hayati-etal-2018-retrieval}
Shirley~Anugrah Hayati, Raphael Olivier, Pravalika Avvaru, Pengcheng Yin,
  Anthony Tomasic, and Graham Neubig. 2018.
\newblock \href {https://doi.org/10.18653/v1/D18-1111} {Retrieval-based neural
  code generation}.
\newblock In \emph{Proceedings of the 2018 Conference on Empirical Methods in
  Natural Language Processing}, pages 925--930, Brussels, Belgium. Association
  for Computational Linguistics.

\bibitem[{He et~al.(2022{\natexlab{a}})He, Zhang, and Roth}]{he2022rethinking}
Hangfeng He, Hongming Zhang, and Dan Roth. 2022{\natexlab{a}}.
\newblock Rethinking with retrieval: Faithful large language model inference.
\newblock \emph{arXiv preprint arXiv:2301.00303}.

\bibitem[{He et~al.(2021)He, Huang, Cui, Li, and Liu}]{he2021fast}
Qiuxiang He, Guoping Huang, Qu~Cui, Li~Li, and Lemao Liu. 2021.
\newblock Fast and accurate neural machine translation with translation memory.
\newblock In \emph{Proceedings of the 59th Annual Meeting of the Association
  for Computational Linguistics and the 11th International Joint Conference on
  Natural Language Processing (Volume 1: Long Papers)}, pages 3170--3180.

\bibitem[{He et~al.(2022{\natexlab{b}})He, Hao, and Song}]{he2022recap}
Zihao He, Weituo Hao, and Xuchen Song. 2022{\natexlab{b}}.
\newblock Recap: Retrieval augmented music captioner.
\newblock \emph{arXiv preprint arXiv:2212.10901}.

\bibitem[{Hu et~al.(2022{\natexlab{a}})Hu, Xiang, Qin, and Tan}]{hu2022audio}
Tao Hu, Xuyu Xiang, Jiaohua Qin, and Yun Tan. 2022{\natexlab{a}}.
\newblock Audio-text retrieval based on contrastive learning and collaborative
  attention mechanism.

\bibitem[{Hu et~al.(2022{\natexlab{b}})Hu, Wu, Shu, and
  Qu}]{hu-etal-2022-logical}
Xixin Hu, Xuan Wu, Yiheng Shu, and Yuzhong Qu. 2022{\natexlab{b}}.
\newblock \href {https://aclanthology.org/2022.coling-1.145} {Logical form
  generation via multi-task learning for complex question answering over
  knowledge bases}.
\newblock In \emph{Proceedings of the 29th International Conference on
  Computational Linguistics}, pages 1687--1696, Gyeongju, Republic of Korea.
  International Committee on Computational Linguistics.

\bibitem[{Hu et~al.(2023)Hu, Iscen, Sun, Wang, Chang, Sun, Schmid, Ross, and
  Fathi}]{hu2023reveal}
Ziniu Hu, Ahmet Iscen, Chen Sun, Zirui Wang, Kai-Wei Chang, Yizhou Sun,
  Cordelia Schmid, David~A Ross, and Alireza Fathi. 2023.
\newblock Reveal: Retrieval-augmented visual-language pre-training with
  multi-source multimodal knowledge memory.
\newblock In \emph{Proceedings of the IEEE/CVF Conference on Computer Vision
  and Pattern Recognition}, pages 23369--23379.

\bibitem[{Hu et~al.(2022{\natexlab{c}})Hu, Xu, Yu, Wang, Yang, Zhu, Chang, and
  Sun}]{hu-etal-2022-empowering}
Ziniu Hu, Yichong Xu, Wenhao Yu, Shuohang Wang, Ziyi Yang, Chenguang Zhu,
  Kai-Wei Chang, and Yizhou Sun. 2022{\natexlab{c}}.
\newblock \href {https://aclanthology.org/2022.emnlp-main.650} {Empowering
  language models with knowledge graph reasoning for open-domain question
  answering}.
\newblock In \emph{Proceedings of the 2022 Conference on Empirical Methods in
  Natural Language Processing}, pages 9562--9581, Abu Dhabi, United Arab
  Emirates. Association for Computational Linguistics.

\bibitem[{Huang et~al.(2023{\natexlab{a}})Huang, Huang, Yang, Ren, Liu, Li, Ye,
  Liu, Yin, and Zhao}]{huang2023make}
Rongjie Huang, Jiawei Huang, Dongchao Yang, Yi~Ren, Luping Liu, Mingze Li,
  Zhenhui Ye, Jinglin Liu, Xiang Yin, and Zhou Zhao. 2023{\natexlab{a}}.
\newblock Make-an-audio: Text-to-audio generation with prompt-enhanced
  diffusion models.
\newblock \emph{arXiv preprint arXiv:2301.12661}.

\bibitem[{Huang et~al.(2023{\natexlab{b}})Huang, Dong, Wang, Hao, Singhal, Ma,
  Lv, Cui, Mohammed, Liu et~al.}]{huang2023language}
Shaohan Huang, Li~Dong, Wenhui Wang, Yaru Hao, Saksham Singhal, Shuming Ma,
  Tengchao Lv, Lei Cui, Owais~Khan Mohammed, Qiang Liu, et~al.
  2023{\natexlab{b}}.
\newblock Language is not all you need: Aligning perception with language
  models.
\newblock \emph{arXiv preprint arXiv:2302.14045}.

\bibitem[{Izacard and Grave(2021)}]{izacard-grave-2021-leveraging}
Gautier Izacard and Edouard Grave. 2021.
\newblock \href {https://doi.org/10.18653/v1/2021.eacl-main.74} {Leveraging
  passage retrieval with generative models for open domain question answering}.
\newblock In \emph{Proceedings of the 16th Conference of the European Chapter
  of the Association for Computational Linguistics: Main Volume}, pages
  874--880, Online. Association for Computational Linguistics.

\bibitem[{Izacard et~al.(2022)Izacard, Lewis, Lomeli, Hosseini, Petroni,
  Schick, Dwivedi-Yu, Joulin, Riedel, and Grave}]{izacard2022atlas}
Gautier Izacard, Patrick Lewis, Maria Lomeli, Lucas Hosseini, Fabio Petroni,
  Timo Schick, Jane Dwivedi-Yu, Armand Joulin, Sebastian Riedel, and Edouard
  Grave. 2022.
\newblock Atlas: Few-shot learning with retrieval augmented language models.
\newblock \emph{arXiv preprint arXiv}, 2208.

\bibitem[{Jansen and Ustalov(2019)}]{jansen-ustalov-2019-textgraphs}
Peter Jansen and Dmitry Ustalov. 2019.
\newblock \href {https://doi.org/10.18653/v1/D19-5309} {{T}ext{G}raphs 2019
  shared task on multi-hop inference for explanation regeneration}.
\newblock In \emph{Proceedings of the Thirteenth Workshop on Graph-Based
  Methods for Natural Language Processing (TextGraphs-13)}, pages 63--77, Hong
  Kong. Association for Computational Linguistics.

\bibitem[{Jia et~al.(2021)Jia, Yang, Xia, Chen, Parekh, Pham, Le, Sung, Li, and
  Duerig}]{jia2021scaling}
Chao Jia, Yinfei Yang, Ye~Xia, Yi-Ting Chen, Zarana Parekh, Hieu Pham, Quoc Le,
  Yun-Hsuan Sung, Zhen Li, and Tom Duerig. 2021.
\newblock Scaling up visual and vision-language representation learning with
  noisy text supervision.
\newblock In \emph{International Conference on Machine Learning}, pages
  4904--4916. PMLR.

\bibitem[{Jiang et~al.(2018)Jiang, Xiong, Zhang, Gao, and Chen}]{simfix}
Jiajun Jiang, Yingfei Xiong, Hongyu Zhang, Qing Gao, and Xiangqun Chen. 2018.
\newblock Shaping program repair space with existing patches and similar code.
\newblock In \emph{{ISSTA}}, pages 298--309. {ACM}.

\bibitem[{Jin et~al.(2022)Jin, Zhang, Zhu, and Han}]{jin2022heterformer}
Bowen Jin, Yu~Zhang, Qi~Zhu, and Jiawei Han. 2022.
\newblock Heterformer: A transformer architecture for node representation
  learning on heterogeneous text-rich networks.
\newblock \emph{arXiv preprint arXiv:2205.10282}.

\bibitem[{Jin et~al.(2023)Jin, Shahriar, Tufano, Shi, Lu, Sundaresan, and
  Svyatkovskiy}]{jin2023inferfix}
Matthew Jin, Syed Shahriar, Michele Tufano, Xin Shi, Shuai Lu, Neel Sundaresan,
  and Alexey Svyatkovskiy. 2023.
\newblock Inferfix: End-to-end program repair with llms.
\newblock \emph{arXiv preprint arXiv:2303.07263}.

\bibitem[{Jo et~al.(2021)Jo, Yoo, Bak, Oh, Reed, and
  Hovy}]{jo-etal-2021-knowledge-enhanced}
Yohan Jo, Haneul Yoo, JinYeong Bak, Alice Oh, Chris Reed, and Eduard Hovy.
  2021.
\newblock \href {https://doi.org/10.18653/v1/2021.findings-emnlp.264}
  {Knowledge-enhanced evidence retrieval for counterargument generation}.
\newblock In \emph{Findings of the Association for Computational Linguistics:
  EMNLP 2021}, pages 3074--3094, Punta Cana, Dominican Republic. Association
  for Computational Linguistics.

\bibitem[{Joshi et~al.(2022)Joshi, Cambronero, Gulwani, Le, Radicek, and
  Verbruggen}]{joshi2022repair}
Harshit Joshi, Jos{\'e} Cambronero, Sumit Gulwani, Vu~Le, Ivan Radicek, and
  Gust Verbruggen. 2022.
\newblock Repair is nearly generation: Multilingual program repair with llms.
\newblock \emph{arXiv preprint arXiv:2208.11640}.

\bibitem[{Ju et~al.(2022)Ju, Han, Zheng, Zhang, and Xie}]{ju2022prompting}
Chen Ju, Tengda Han, Kunhao Zheng, Ya~Zhang, and Weidi Xie. 2022.
\newblock Prompting visual-language models for efficient video understanding.
\newblock In \emph{Computer Vision--ECCV 2022: 17th European Conference, Tel
  Aviv, Israel, October 23--27, 2022, Proceedings, Part XXXV}, pages 105--124.
  Springer.

\bibitem[{Jung et~al.(2020)Jung, Son, and Lyu}]{jung-etal-2020-attnio}
Jaehun Jung, Bokyung Son, and Sungwon Lyu. 2020.
\newblock \href {https://doi.org/10.18653/v1/2020.emnlp-main.280} {{A}ttn{IO}:
  {K}nowledge {G}raph {E}xploration with {I}n-and-{O}ut {A}ttention {F}low for
  {K}nowledge-{G}rounded {D}ialogue}.
\newblock In \emph{Proceedings of the 2020 Conference on Empirical Methods in
  Natural Language Processing (EMNLP)}, pages 3484--3497, Online. Association
  for Computational Linguistics.

\bibitem[{Kang et~al.(2023)Kang, Kwak, Baek, and Hwang}]{kang2023knowledge}
Minki Kang, Jin~Myung Kwak, Jinheon Baek, and Sung~Ju Hwang. 2023.
\newblock \href {http://arxiv.org/abs/2305.18846} {Knowledge graph-augmented
  language models for knowledge-grounded dialogue generation}.

\bibitem[{Karpukhin et~al.(2020{\natexlab{a}})Karpukhin, O{\u{g}}uz, Min,
  Lewis, Wu, Edunov, Chen, and Yih}]{dpr}
Vladimir Karpukhin, Barlas O{\u{g}}uz, Sewon Min, Patrick Lewis, Ledell Wu,
  Sergey Edunov, Danqi Chen, and Wen-tau Yih. 2020{\natexlab{a}}.
\newblock Dense passage retrieval for open-domain question answering.
\newblock \emph{arXiv preprint arXiv:2004.04906}.

\bibitem[{Karpukhin et~al.(2020{\natexlab{b}})Karpukhin, Oguz, Min, Lewis, Wu,
  Edunov, Chen, and Yih}]{karpukhin-etal-2020-dense}
Vladimir Karpukhin, Barlas Oguz, Sewon Min, Patrick Lewis, Ledell Wu, Sergey
  Edunov, Danqi Chen, and Wen-tau Yih. 2020{\natexlab{b}}.
\newblock \href {https://doi.org/10.18653/v1/2020.emnlp-main.550} {Dense
  passage retrieval for open-domain question answering}.
\newblock In \emph{Proceedings of the 2020 Conference on Empirical Methods in
  Natural Language Processing (EMNLP)}, pages 6769--6781, Online. Association
  for Computational Linguistics.

\bibitem[{Kim et~al.(2023)Kim, Sung-Bin, and Oh}]{kim2023prefix}
Minkyu Kim, Kim Sung-Bin, and Tae-Hyun Oh. 2023.
\newblock Prefix tuning for automated audio captioning.
\newblock In \emph{ICASSP 2023-2023 IEEE International Conference on Acoustics,
  Speech and Signal Processing (ICASSP)}, pages 1--5. IEEE.

\bibitem[{Koepke et~al.(2022)Koepke, Oncescu, Henriques, Akata, and
  Albanie}]{koepke2022audio}
A~Sophia Koepke, Andreea-Maria Oncescu, Joao Henriques, Zeynep Akata, and
  Samuel Albanie. 2022.
\newblock Audio retrieval with natural language queries: A benchmark study.
\newblock \emph{IEEE Transactions on Multimedia}.

\bibitem[{Koizumi et~al.(2020)Koizumi, Ohishi, Niizumi, Takeuchi, and
  Yasuda}]{koizumi2020audio}
Yuma Koizumi, Yasunori Ohishi, Daisuke Niizumi, Daiki Takeuchi, and Masahiro
  Yasuda. 2020.
\newblock Audio captioning using pre-trained large-scale language model guided
  by audio-based similar caption retrieval.
\newblock \emph{arXiv preprint arXiv:2012.07331}.

\bibitem[{Le et~al.(2022)Le, Chen, and Hoi}]{le2022vgnmn}
Hung Le, Nancy Chen, and Steven Hoi. 2022.
\newblock Vgnmn: Video-grounded neural module networks for video-grounded
  dialogue systems.
\newblock In \emph{Proceedings of the 2022 Conference of the North American
  Chapter of the Association for Computational Linguistics: Human Language
  Technologies}, pages 3377--3393.

\bibitem[{Le et~al.(2020)Le, Sahoo, Chen, and Hoi}]{le-etal-2020-bist}
Hung Le, Doyen Sahoo, Nancy Chen, and Steven~C.H. Hoi. 2020.
\newblock \href {https://doi.org/10.18653/v1/2020.emnlp-main.145} {{B}i{ST}:
  Bi-directional spatio-temporal reasoning for video-grounded dialogues}.
\newblock In \emph{Proceedings of the 2020 Conference on Empirical Methods in
  Natural Language Processing (EMNLP)}, pages 1846--1859, Online. Association
  for Computational Linguistics.

\bibitem[{Lee et~al.(2021{\natexlab{a}})Lee, Sung, Kang, and
  Chen}]{lee-etal-2021-learning-dense}
Jinhyuk Lee, Mujeen Sung, Jaewoo Kang, and Danqi Chen. 2021{\natexlab{a}}.
\newblock \href {https://doi.org/10.18653/v1/2021.acl-long.518} {Learning dense
  representations of phrases at scale}.
\newblock In \emph{Proceedings of the 59th Annual Meeting of the Association
  for Computational Linguistics and the 11th International Joint Conference on
  Natural Language Processing (Volume 1: Long Papers)}, pages 6634--6647,
  Online. Association for Computational Linguistics.

\bibitem[{Lee et~al.(2021{\natexlab{b}})Lee, Shin, Choo, Choi, and
  Myaeng}]{lee-etal-2021-constructing}
Nyoungwoo Lee, Suwon Shin, Jaegul Choo, Ho-Jin Choi, and Sung-Hyon Myaeng.
  2021{\natexlab{b}}.
\newblock \href {https://doi.org/10.18653/v1/2021.acl-short.113} {Constructing
  multi-modal dialogue dataset by replacing text with semantically relevant
  images}.
\newblock In \emph{Proceedings of the 59th Annual Meeting of the Association
  for Computational Linguistics and the 11th International Joint Conference on
  Natural Language Processing (Volume 2: Short Papers)}, pages 897--906,
  Online. Association for Computational Linguistics.

\bibitem[{Lei et~al.(2021)Lei, Li, Zhou, Gan, Berg, Bansal, and
  Liu}]{lei2021less}
Jie Lei, Linjie Li, Luowei Zhou, Zhe Gan, Tamara~L Berg, Mohit Bansal, and
  Jingjing Liu. 2021.
\newblock Less is more: Clipbert for video-and-language learning via sparse
  sampling.
\newblock In \emph{Proceedings of the IEEE/CVF Conference on Computer Vision
  and Pattern Recognition}, pages 7331--7341.

\bibitem[{Lei et~al.(2020)Lei, Yu, Berg, and Bansal}]{lei-etal-2020-tvqa}
Jie Lei, Licheng Yu, Tamara Berg, and Mohit Bansal. 2020.
\newblock \href {https://doi.org/10.18653/v1/2020.acl-main.730} {{TVQA}+:
  Spatio-temporal grounding for video question answering}.
\newblock In \emph{Proceedings of the 58th Annual Meeting of the Association
  for Computational Linguistics}, pages 8211--8225, Online. Association for
  Computational Linguistics.

\bibitem[{Leskovec et~al.(2014)Leskovec, Rajaraman, and Ullman}]{data-mining}
Jure Leskovec, Anand Rajaraman, and Jeffrey~D. Ullman. 2014.
\newblock \href {http://www.mmds.org/} {\emph{Mining of Massive Datasets, 2nd
  Ed}}.
\newblock Cambridge University Press.

\bibitem[{Lewis et~al.(2020)Lewis, Perez, Piktus, Petroni, Karpukhin, Goyal,
  K{\"u}ttler, Lewis, Yih, Rockt{\"a}schel et~al.}]{lewis2020retrieval}
Patrick Lewis, Ethan Perez, Aleksandra Piktus, Fabio Petroni, Vladimir
  Karpukhin, Naman Goyal, Heinrich K{\"u}ttler, Mike Lewis, Wen-tau Yih, Tim
  Rockt{\"a}schel, et~al. 2020.
\newblock Retrieval-augmented generation for knowledge-intensive nlp tasks.
\newblock \emph{Advances in Neural Information Processing Systems},
  33:9459--9474.

\bibitem[{Li et~al.(2019)Li, Liang, Hu, and Xing}]{li2019knowledgedriven}
Christy~Y. Li, Xiaodan Liang, Zhiting Hu, and Eric~P. Xing. 2019.
\newblock \href {http://arxiv.org/abs/1903.10122} {Knowledge-driven encode,
  retrieve, paraphrase for medical image report generation}.

\bibitem[{Li et~al.(2022{\natexlab{a}})Li, Su, Cai, Wang, and
  Liu}]{li2022survey}
Huayang Li, Yixuan Su, Deng Cai, Yan Wang, and Lemao Liu. 2022{\natexlab{a}}.
\newblock A survey on retrieval-augmented text generation.
\newblock \emph{arXiv preprint arXiv:2202.01110}.

\bibitem[{Li et~al.(2023{\natexlab{a}})Li, Li, Savarese, and Hoi}]{li2023blip}
Junnan Li, Dongxu Li, Silvio Savarese, and Steven Hoi. 2023{\natexlab{a}}.
\newblock Blip-2: Bootstrapping language-image pre-training with frozen image
  encoders and large language models.
\newblock \emph{arXiv preprint arXiv:2301.12597}.

\bibitem[{Li et~al.(2022{\natexlab{b}})Li, Peng, Gao, and Zhang}]{li2022opera}
Miaoran Li, Baolin Peng, Jianfeng Gao, and Zhu Zhang. 2022{\natexlab{b}}.
\newblock Opera: Harmonizing task-oriented dialogs and information seeking
  experience.
\newblock \emph{arXiv preprint arXiv:2206.12449}.

\bibitem[{Li et~al.(2023{\natexlab{b}})Li, Zhao, Chia, Ding, Bing, Joty, and
  Poria}]{li2023chain}
Xingxuan Li, Ruochen Zhao, Yew~Ken Chia, Bosheng Ding, Lidong Bing, Shafiq
  Joty, and Soujanya Poria. 2023{\natexlab{b}}.
\newblock \href {http://arxiv.org/abs/2305.13269} {Chain of knowledge: A
  framework for grounding large language models with structured knowledge
  bases}.

\bibitem[{Li et~al.(2022{\natexlab{c}})Li, Peng, Shen, Mao, Liden, Yu, and
  Gao}]{li-etal-2022-knowledge}
Yu~Li, Baolin Peng, Yelong Shen, Yi~Mao, Lars Liden, Zhou Yu, and Jianfeng Gao.
  2022{\natexlab{c}}.
\newblock \href {https://doi.org/10.18653/v1/2022.naacl-main.15}
  {Knowledge-grounded dialogue generation with a unified knowledge
  representation}.
\newblock In \emph{Proceedings of the 2022 Conference of the North American
  Chapter of the Association for Computational Linguistics: Human Language
  Technologies}, pages 206--218, Seattle, United States. Association for
  Computational Linguistics.

\bibitem[{Li et~al.(2022{\natexlab{d}})Li, Lu, Guo, Duan, Jannu, Jenks,
  Majumder, Green, Svyatkovskiy, Fu, and Sundaresan}]{codereviewer}
Zhiyu Li, Shuai Lu, Daya Guo, Nan Duan, Shailesh Jannu, Grant Jenks, Deep
  Majumder, Jared Green, Alexey Svyatkovskiy, Shengyu Fu, and Neel Sundaresan.
  2022{\natexlab{d}}.
\newblock \href {https://doi.org/10.1145/3540250.3549081} {Automating code
  review activities by large-scale pre-training}.
\newblock In \emph{Proceedings of the 30th {ACM} Joint European Software
  Engineering Conference and Symposium on the Foundations of Software
  Engineering, {ESEC/FSE} 2022, Singapore, Singapore, November 14-18, 2022},
  pages 1035--1047. {ACM}.

\bibitem[{Liang et~al.(2021)Liang, Hu, Xu, Tao, Geng, Chen, Liang, and
  Jiang}]{liang-etal-2021-maria}
Zujie Liang, Huang Hu, Can Xu, Chongyang Tao, Xiubo Geng, Yining Chen, Fan
  Liang, and Daxin Jiang. 2021.
\newblock \href {https://doi.org/10.18653/v1/2021.acl-long.435} {Maria: A
  visual experience powered conversational agent}.
\newblock In \emph{Proceedings of the 59th Annual Meeting of the Association
  for Computational Linguistics and the 11th International Joint Conference on
  Natural Language Processing (Volume 1: Long Papers)}, pages 5596--5611,
  Online. Association for Computational Linguistics.

\bibitem[{Lin and Byrne(2022{\natexlab{a}})}]{RA-VQA}
Weizhe Lin and Bill Byrne. 2022{\natexlab{a}}.
\newblock Retrieval augmented visual question answering with outside knowledge.
\newblock In \emph{{EMNLP}}, pages 11238--11254. Association for Computational
  Linguistics.

\bibitem[{Lin and Byrne(2022{\natexlab{b}})}]{lin-byrne-2022-retrieval}
Weizhe Lin and Bill Byrne. 2022{\natexlab{b}}.
\newblock \href {https://aclanthology.org/2022.emnlp-main.772} {Retrieval
  augmented visual question answering with outside knowledge}.
\newblock In \emph{Proceedings of the 2022 Conference on Empirical Methods in
  Natural Language Processing}, pages 11238--11254, Abu Dhabi, United Arab
  Emirates. Association for Computational Linguistics.

\bibitem[{Lin et~al.(2023)Lin, Wang, and Byrne}]{lin-etal-2023-fvqa}
Weizhe Lin, Zhilin Wang, and Bill Byrne. 2023.
\newblock \href {https://aclanthology.org/2023.findings-eacl.11} {{FVQA} 2.0:
  Introducing adversarial samples into fact-based visual question answering}.
\newblock In \emph{Findings of the Association for Computational Linguistics:
  EACL 2023}, pages 149--157, Dubrovnik, Croatia. Association for Computational
  Linguistics.

\bibitem[{Liu et~al.(2022{\natexlab{a}})Liu, Yogatama, and
  Blunsom}]{liu-etal-2022-relational}
Qi~Liu, Dani Yogatama, and Phil Blunsom. 2022{\natexlab{a}}.
\newblock \href {https://doi.org/10.1162/tacl_a_00476} {Relational
  memory-augmented language models}.
\newblock \emph{Transactions of the Association for Computational Linguistics},
  10:555--572.

\bibitem[{Liu et~al.(2021)Liu, Chen, Xie, Siow, and
  Liu}]{liu2021retrievalaugmented}
Shangqing Liu, Yu~Chen, Xiaofei Xie, Jing~Kai Siow, and Yang Liu. 2021.
\newblock \href {https://openreview.net/forum?id=zv-typ1gPxA}
  {Retrieval-augmented generation for code summarization via hybrid {GNN}}.
\newblock In \emph{{ICLR}}.

\bibitem[{Liu et~al.(2015)Liu, Chen, Chen, Wang, Yen, and
  Hsu}]{liu2015combining}
Shih-Hung Liu, Kuan-Yu Chen, Berlin Chen, Hsin-Min Wang, Hsu-Chun Yen, and
  Wen-Lian Hsu. 2015.
\newblock Combining relevance language modeling and clarity measure for
  extractive speech summarization.
\newblock \emph{IEEE/ACM Transactions on Audio, Speech, and Language
  Processing}, 23(6):957--969.

\bibitem[{Liu and Zhong(2018)}]{DBLP:conf/wcre/LiuZ18}
Xuliang Liu and Hao Zhong. 2018.
\newblock Mining stackoverflow for program repair.
\newblock In \emph{{SANER}}, pages 118--129. {IEEE} Computer Society.

\bibitem[{Liu et~al.(2022{\natexlab{b}})Liu, Yavuz, Meng, Radev, Xiong, and
  Zhou}]{liu-etal-2022-uni}
Ye~Liu, Semih Yavuz, Rui Meng, Dragomir Radev, Caiming Xiong, and Yingbo Zhou.
  2022{\natexlab{b}}.
\newblock \href {https://aclanthology.org/2022.emnlp-main.605} {Uni-parser:
  Unified semantic parser for question answering on knowledge base and
  database}.
\newblock In \emph{Proceedings of the 2022 Conference on Empirical Methods in
  Natural Language Processing}, pages 8858--8869, Abu Dhabi, United Arab
  Emirates. Association for Computational Linguistics.

\bibitem[{Long et~al.(2018)Long, Gan, and De~Melo}]{long2018video}
Xiang Long, Chuang Gan, and Gerard De~Melo. 2018.
\newblock Video captioning with multi-faceted attention.
\newblock \emph{Transactions of the Association for Computational Linguistics},
  6:173--184.

\bibitem[{Lou et~al.(2022)Lou, Xu, Wu, and Yu}]{lou2022audio}
Siyu Lou, Xuenan Xu, Mengyue Wu, and Kai Yu. 2022.
\newblock Audio-text retrieval in context.
\newblock In \emph{ICASSP 2022-2022 IEEE International Conference on Acoustics,
  Speech and Signal Processing (ICASSP)}, pages 4793--4797. IEEE.

\bibitem[{Lu et~al.(2022{\natexlab{a}})Lu, Mishra, Xia, Qiu, Chang, Zhu,
  Tafjord, Clark, and Kalyan}]{lu2022learn}
Pan Lu, Swaroop Mishra, Tony Xia, Liang Qiu, Kai-Wei Chang, Song-Chun Zhu,
  Oyvind Tafjord, Peter Clark, and Ashwin Kalyan. 2022{\natexlab{a}}.
\newblock Learn to explain: Multimodal reasoning via thought chains for science
  question answering.
\newblock \emph{arXiv preprint arXiv:2209.09513}.

\bibitem[{Lu et~al.(2022{\natexlab{b}})Lu, Duan, Han, Guo, Hwang, and
  Svyatkovskiy}]{lu-etal-2022-reacc}
Shuai Lu, Nan Duan, Hojae Han, Daya Guo, Seung-won Hwang, and Alexey
  Svyatkovskiy. 2022{\natexlab{b}}.
\newblock \href {https://doi.org/10.18653/v1/2022.acl-long.431} {{R}e{ACC}: A
  retrieval-augmented code completion framework}.
\newblock In \emph{Proceedings of the 60th Annual Meeting of the Association
  for Computational Linguistics (Volume 1: Long Papers)}, pages 6227--6240,
  Dublin, Ireland. Association for Computational Linguistics.

\bibitem[{Lu et~al.(2023)Lu, Reddy, and de~Vries}]{lu-etal-2023-statcan}
Xing~Han Lu, Siva Reddy, and Harm de~Vries. 2023.
\newblock \href {https://aclanthology.org/2023.eacl-main.206} {The {S}tat{C}an
  dialogue dataset: Retrieving data tables through conversations with genuine
  intents}.
\newblock In \emph{Proceedings of the 17th Conference of the European Chapter
  of the Association for Computational Linguistics}, pages 2799--2829,
  Dubrovnik, Croatia. Association for Computational Linguistics.

\bibitem[{Lyu et~al.(2023)Lyu, Havaldar, Stein, Zhang, Rao, Wong, Apidianaki,
  and Callison-Burch}]{lyu2023faithful}
Qing Lyu, Shreya Havaldar, Adam Stein, Li~Zhang, Delip Rao, Eric Wong, Marianna
  Apidianaki, and Chris Callison-Burch. 2023.
\newblock Faithful chain-of-thought reasoning.
\newblock \emph{arXiv preprint arXiv:2301.13379}.

\bibitem[{Ma et~al.(2022)Ma, Cheng, Liu, Nyberg, and
  Gao}]{ma-etal-2022-open-domain}
Kaixin Ma, Hao Cheng, Xiaodong Liu, Eric Nyberg, and Jianfeng Gao. 2022.
\newblock \href {https://aclanthology.org/2022.findings-emnlp.392} {Open-domain
  question answering via chain of reasoning over heterogeneous knowledge}.
\newblock In \emph{Findings of the Association for Computational Linguistics:
  EMNLP 2022}, pages 5360--5374, Abu Dhabi, United Arab Emirates. Association
  for Computational Linguistics.

\bibitem[{Madaan et~al.(2022)Madaan, Zhou, Alon, Yang, and
  Neubig}]{madaan2022language}
Aman Madaan, Shuyan Zhou, Uri Alon, Yiming Yang, and Graham Neubig. 2022.
\newblock Language models of code are few-shot commonsense learners.
\newblock \emph{arXiv preprint arXiv:2210.07128}.

\bibitem[{Manco et~al.(2021)Manco, Benetos, Quinton, and
  Fazekas}]{manco2021muscaps}
Ilaria Manco, Emmanouil Benetos, Elio Quinton, and Gy{\"o}rgy Fazekas. 2021.
\newblock Muscaps: Generating captions for music audio.
\newblock In \emph{2021 International Joint Conference on Neural Networks
  (IJCNN)}, pages 1--8. IEEE.

\bibitem[{Martinez et~al.(2014)Martinez, Weimer, and Martin}]{Martinez2014DoTF}
Matias Martinez, Westley Weimer, and Monperrus Martin. 2014.
\newblock Do the fix ingredients already exist? an empirical inquiry into the
  redundancy assumptions of program repair approaches.
\newblock \emph{Companion Proceedings of the 36th International Conference on
  Software Engineering}.

\bibitem[{McConnell(2004)}]{mcconnell2004code}
Steve McConnell. 2004.
\newblock \emph{Code complete}.
\newblock Pearson Education.

\bibitem[{Mestre et~al.(2023)Mestre, Middleton, Ryan, Gheasi, Norman, and
  Zhu}]{mestre-etal-2023-augmenting}
Rafael Mestre, Stuart Middleton, Matt Ryan, Masood Gheasi, Timothy Norman, and
  Jiatong Zhu. 2023.
\newblock \href {https://aclanthology.org/2023.findings-eacl.21} {Augmenting
  pre-trained language models with audio feature embedding for argumentation
  mining in political debates}.
\newblock In \emph{Findings of the Association for Computational Linguistics:
  EACL 2023}, pages 274--288, Dubrovnik, Croatia. Association for Computational
  Linguistics.

\bibitem[{Mialon et~al.(2023)Mialon, Dess{\`\i}, Lomeli, Nalmpantis, Pasunuru,
  Raileanu, Rozi{\`e}re, Schick, Dwivedi-Yu, Celikyilmaz
  et~al.}]{mialon2023augmented}
Gr{\'e}goire Mialon, Roberto Dess{\`\i}, Maria Lomeli, Christoforos Nalmpantis,
  Ram Pasunuru, Roberta Raileanu, Baptiste Rozi{\`e}re, Timo Schick, Jane
  Dwivedi-Yu, Asli Celikyilmaz, et~al. 2023.
\newblock Augmented language models: a survey.
\newblock \emph{arXiv preprint arXiv:2302.07842}.

\bibitem[{Milde et~al.(2016{\natexlab{a}})Milde, Wacker, Radomski,
  M{\"u}hlh{\"a}user, and Biemann}]{milde-etal-2016-ambient}
Benjamin Milde, Jonas Wacker, Stefan Radomski, Max M{\"u}hlh{\"a}user, and
  Chris Biemann. 2016{\natexlab{a}}.
\newblock \href {https://aclanthology.org/C16-1196} {Ambient search: A document
  retrieval system for speech streams}.
\newblock In \emph{Proceedings of {COLING} 2016, the 26th International
  Conference on Computational Linguistics: Technical Papers}, pages 2082--2091,
  Osaka, Japan. The COLING 2016 Organizing Committee.

\bibitem[{Milde et~al.(2016{\natexlab{b}})Milde, Wacker, Radomski,
  M{\"u}hlh{\"a}user, and Biemann}]{milde-etal-2016-demonstrating}
Benjamin Milde, Jonas Wacker, Stefan Radomski, Max M{\"u}hlh{\"a}user, and
  Chris Biemann. 2016{\natexlab{b}}.
\newblock \href {https://aclanthology.org/C16-2049} {Demonstrating ambient
  search: Implicit document retrieval for speech streams}.
\newblock In \emph{Proceedings of {COLING} 2016, the 26th International
  Conference on Computational Linguistics: System Demonstrations}, pages
  233--237, Osaka, Japan. The COLING 2016 Organizing Committee.

\bibitem[{Mille et~al.(2020)Mille, Symeonidis, Rousi, Marimon~Felipe,
  Stavrothanasopoulos, Alvanitopoulos, Carlini~Salguero, Grivolla, Meditskos,
  Vrochidis, and Wanner}]{mille-etal-2020-case}
Simon Mille, Spyridon Symeonidis, Maria Rousi, Montserrat Marimon~Felipe,
  Klearchos Stavrothanasopoulos, Petros Alvanitopoulos, Roberto
  Carlini~Salguero, Jens Grivolla, Georgios Meditskos, Stefanos Vrochidis, and
  Leo Wanner. 2020.
\newblock \href {https://aclanthology.org/2020.webnlg-1.1} {A case study of
  {NLG} from multimedia data sources: Generating architectural landmark
  descriptions}.
\newblock In \emph{Proceedings of the 3rd International Workshop on Natural
  Language Generation from the Semantic Web (WebNLG+)}, pages 2--14, Dublin,
  Ireland (Virtual). Association for Computational Linguistics.

\bibitem[{Mithun et~al.(2018)Mithun, Li, Metze, and
  Roy-Chowdhury}]{mithun2018learning}
Niluthpol~Chowdhury Mithun, Juncheng Li, Florian Metze, and Amit~K
  Roy-Chowdhury. 2018.
\newblock Learning joint embedding with multimodal cues for cross-modal
  video-text retrieval.
\newblock In \emph{Proceedings of the 2018 ACM on International Conference on
  Multimedia Retrieval}, pages 19--27.

\bibitem[{Murugesan et~al.(2021)Murugesan, Atzeni, Kapanipathi, Talamadupula,
  Sachan, and Campbell}]{murugesan-etal-2021-efficient}
Keerthiram Murugesan, Mattia Atzeni, Pavan Kapanipathi, Kartik Talamadupula,
  Mrinmaya Sachan, and Murray Campbell. 2021.
\newblock \href {https://doi.org/10.18653/v1/2021.acl-short.91} {Efficient
  text-based reinforcement learning by jointly leveraging state and commonsense
  graph representations}.
\newblock In \emph{Proceedings of the 59th Annual Meeting of the Association
  for Computational Linguistics and the 11th International Joint Conference on
  Natural Language Processing (Volume 2: Short Papers)}, pages 719--725,
  Online. Association for Computational Linguistics.

\bibitem[{Nagrani et~al.(2021)Nagrani, Yang, Arnab, Jansen, Schmid, and
  Sun}]{nagrani2021attention}
Arsha Nagrani, Shan Yang, Anurag Arnab, Aren Jansen, Cordelia Schmid, and Chen
  Sun. 2021.
\newblock Attention bottlenecks for multimodal fusion.
\newblock \emph{Advances in Neural Information Processing Systems},
  34:14200--14213.

\bibitem[{Nakamura et~al.(2022)Nakamura, Levy, Tuan, Chen, and
  Wang}]{nakamura-etal-2022-hybridialogue}
Kai Nakamura, Sharon Levy, Yi-Lin Tuan, Wenhu Chen, and William~Yang Wang.
  2022.
\newblock \href {https://doi.org/10.18653/v1/2022.findings-acl.41}
  {{H}ybri{D}ialogue: An information-seeking dialogue dataset grounded on
  tabular and textual data}.
\newblock In \emph{Findings of the Association for Computational Linguistics:
  ACL 2022}, pages 481--492, Dublin, Ireland. Association for Computational
  Linguistics.

\bibitem[{Nakano et~al.(2021)Nakano, Hilton, Balaji, Wu, Ouyang, Kim, Hesse,
  Jain, Kosaraju, Saunders et~al.}]{nakano2021webgpt}
Reiichiro Nakano, Jacob Hilton, Suchir Balaji, Jeff Wu, Long Ouyang, Christina
  Kim, Christopher Hesse, Shantanu Jain, Vineet Kosaraju, William Saunders,
  et~al. 2021.
\newblock Webgpt: Browser-assisted question-answering with human feedback.
\newblock \emph{arXiv preprint arXiv:2112.09332}.

\bibitem[{Nan et~al.(2022)Nan, Hsieh, Mao, Lin, Verma, Zhang,
  Kry{\'s}ci{\'n}ski, Schoelkopf, Kong, Tang, Mutuma, Rosand, Trindade,
  Bandaru, Cunningham, Xiong, Radev, and Radev}]{nan-etal-2022-fetaqa}
Linyong Nan, Chiachun Hsieh, Ziming Mao, Xi~Victoria Lin, Neha Verma, Rui
  Zhang, Wojciech Kry{\'s}ci{\'n}ski, Hailey Schoelkopf, Riley Kong, Xiangru
  Tang, Mutethia Mutuma, Ben Rosand, Isabel Trindade, Renusree Bandaru, Jacob
  Cunningham, Caiming Xiong, Dragomir Radev, and Dragomir Radev. 2022.
\newblock \href {https://doi.org/10.1162/tacl_a_00446} {{F}e{T}a{QA}: Free-form
  table question answering}.
\newblock \emph{Transactions of the Association for Computational Linguistics},
  10:35--49.

\bibitem[{Nashid et~al.(2023)Nashid, Sintaha, and Mesbah}]{nashidretrieval}
Noor Nashid, Mifta Sintaha, and Ali Mesbah. 2023.
\newblock Retrieval-based prompt selection for code-related few-shot learning.
\newblock In \emph{Proceedings of the 45th International Conference on Software
  Engineering (ICSE’23)}.

\bibitem[{Neves~Ribeiro et~al.(2022)Neves~Ribeiro, Wang, Ma, Dong, Wei, Zhu,
  Chen, Xu, Huang, Arnold, and Roth}]{neves-ribeiro-etal-2022-entailment}
Danilo Neves~Ribeiro, Shen Wang, Xiaofei Ma, Rui Dong, Xiaokai Wei, Henghui
  Zhu, Xinchi Chen, Peng Xu, Zhiheng Huang, Andrew Arnold, and Dan Roth. 2022.
\newblock \href {https://doi.org/10.18653/v1/2022.findings-naacl.35}
  {Entailment tree explanations via iterative retrieval-generation reasoner}.
\newblock In \emph{Findings of the Association for Computational Linguistics:
  NAACL 2022}, pages 465--475, Seattle, United States. Association for
  Computational Linguistics.

\bibitem[{{OpenAI}(2023)}]{gpt4}
{OpenAI}. 2023.
\newblock \href {https://doi.org/10.48550/ARXIV.2303.08774} {Gpt-4 technical
  report}.

\bibitem[{Ouyang et~al.(2022)Ouyang, Wu, Jiang, Almeida, Wainwright, Mishkin,
  Zhang, Agarwal, Slama, Ray et~al.}]{ouyang2022training}
Long Ouyang, Jeff Wu, Xu~Jiang, Diogo Almeida, Carroll~L Wainwright, Pamela
  Mishkin, Chong Zhang, Sandhini Agarwal, Katarina Slama, Alex Ray, et~al.
  2022.
\newblock Training language models to follow instructions with human feedback.
\newblock \emph{arXiv preprint arXiv:2203.02155}.

\bibitem[{Pan et~al.(2021)Pan, Canim, Glass, Gliozzo, and
  Fox}]{pan-etal-2021-cltr}
Feifei Pan, Mustafa Canim, Michael Glass, Alfio Gliozzo, and Peter Fox. 2021.
\newblock \href {https://doi.org/10.18653/v1/2021.acl-demo.24} {{CLTR}: An
  end-to-end, transformer-based system for cell-level table retrieval and table
  question answering}.
\newblock In \emph{Proceedings of the 59th Annual Meeting of the Association
  for Computational Linguistics and the 11th International Joint Conference on
  Natural Language Processing: System Demonstrations}, pages 202--209, Online.
  Association for Computational Linguistics.

\bibitem[{Parvez et~al.(2021)Parvez, Ahmad, Chakraborty, Ray, and
  Chang}]{parvez-etal-2021-retrieval-augmented}
Md~Rizwan Parvez, Wasi Ahmad, Saikat Chakraborty, Baishakhi Ray, and Kai-Wei
  Chang. 2021.
\newblock \href {https://doi.org/10.18653/v1/2021.findings-emnlp.232}
  {Retrieval augmented code generation and summarization}.
\newblock In \emph{Findings of the Association for Computational Linguistics:
  EMNLP 2021}, pages 2719--2734, Punta Cana, Dominican Republic. Association
  for Computational Linguistics.

\bibitem[{Pasunuru and Bansal(2018)}]{pasunuru2018game}
Ramakanth Pasunuru and Mohit Bansal. 2018.
\newblock Game-based video-context dialogue.
\newblock \emph{arXiv preprint arXiv:1809.04560}.

\bibitem[{Patel et~al.(2021)Patel, Bhattamishra, and Goyal}]{math}
Arkil Patel, Satwik Bhattamishra, and Navin Goyal. 2021.
\newblock Are nlp models really able to solve simple math word problems?
\newblock \emph{arXiv preprint arXiv:2103.07191}.

\bibitem[{Peng et~al.(2019)Peng, Parikh, Faruqui, Dhingra, and
  Das}]{peng-etal-2019-text}
Hao Peng, Ankur Parikh, Manaal Faruqui, Bhuwan Dhingra, and Dipanjan Das. 2019.
\newblock \href {https://doi.org/10.18653/v1/N19-1263} {Text generation with
  exemplar-based adaptive decoding}.
\newblock In \emph{Proceedings of the 2019 Conference of the North {A}merican
  Chapter of the Association for Computational Linguistics: Human Language
  Technologies, Volume 1 (Long and Short Papers)}, pages 2555--2565,
  Minneapolis, Minnesota. Association for Computational Linguistics.

\bibitem[{Poole et~al.(2022)Poole, Jain, Barron, and
  Mildenhall}]{poole2022dreamfusion}
Ben Poole, Ajay Jain, Jonathan~T Barron, and Ben Mildenhall. 2022.
\newblock Dreamfusion: Text-to-3d using 2d diffusion.
\newblock \emph{arXiv preprint arXiv:2209.14988}.

\bibitem[{Pramanik et~al.(2021)Pramanik, Alabi, Roy, and
  Weikum}]{pramanik2021uniqorn}
Soumajit Pramanik, Jesujoba Alabi, Rishiraj~Saha Roy, and Gerhard Weikum. 2021.
\newblock Uniqorn: unified question answering over rdf knowledge graphs and
  natural language text.
\newblock \emph{arXiv preprint arXiv:2108.08614}.

\bibitem[{Qi et~al.(2014)Qi, Mao, Lei, Dai, and Wang}]{DBLP:conf/icse/QiMLDW14}
Yuhua Qi, Xiaoguang Mao, Yan Lei, Ziying Dai, and Chengsong Wang. 2014.
\newblock The strength of random search on automated program repair.
\newblock In \emph{{ICSE}}, pages 254--265. {ACM}.

\bibitem[{Radford et~al.(2021{\natexlab{a}})Radford, Kim, Hallacy, Ramesh, Goh,
  Agarwal, Sastry, Askell, Mishkin, Clark, Krueger, and Sutskever}]{clip}
Alec Radford, Jong~Wook Kim, Chris Hallacy, Aditya Ramesh, Gabriel Goh,
  Sandhini Agarwal, Girish Sastry, Amanda Askell, Pamela Mishkin, Jack Clark,
  Gretchen Krueger, and Ilya Sutskever. 2021{\natexlab{a}}.
\newblock \href {http://proceedings.mlr.press/v139/radford21a.html} {Learning
  transferable visual models from natural language supervision}.
\newblock In \emph{Proceedings of the 38th International Conference on Machine
  Learning, {ICML} 2021, 18-24 July 2021, Virtual Event}, volume 139 of
  \emph{Proceedings of Machine Learning Research}, pages 8748--8763. {PMLR}.

\bibitem[{Radford et~al.(2021{\natexlab{b}})Radford, Kim, Hallacy, Ramesh, Goh,
  Agarwal, Sastry, Askell, Mishkin, Clark et~al.}]{radford2021learning}
Alec Radford, Jong~Wook Kim, Chris Hallacy, Aditya Ramesh, Gabriel Goh,
  Sandhini Agarwal, Girish Sastry, Amanda Askell, Pamela Mishkin, Jack Clark,
  et~al. 2021{\natexlab{b}}.
\newblock Learning transferable visual models from natural language
  supervision.
\newblock In \emph{International conference on machine learning}, pages
  8748--8763. PMLR.

\bibitem[{Ramesh et~al.(2021{\natexlab{a}})Ramesh, Pavlov, Goh, Gray, Voss,
  Radford, Chen, and Sutskever}]{DALL-E}
Aditya Ramesh, Mikhail Pavlov, Gabriel Goh, Scott Gray, Chelsea Voss, Alec
  Radford, Mark Chen, and Ilya Sutskever. 2021{\natexlab{a}}.
\newblock Zero-shot text-to-image generation.
\newblock In \emph{{ICML}}, volume 139 of \emph{Proceedings of Machine Learning
  Research}, pages 8821--8831. {PMLR}.

\bibitem[{Ramesh et~al.(2021{\natexlab{b}})Ramesh, Pavlov, Goh, Gray, Voss,
  Radford, Chen, and Sutskever}]{ramesh2021zero}
Aditya Ramesh, Mikhail Pavlov, Gabriel Goh, Scott Gray, Chelsea Voss, Alec
  Radford, Mark Chen, and Ilya Sutskever. 2021{\natexlab{b}}.
\newblock Zero-shot text-to-image generation.
\newblock In \emph{International Conference on Machine Learning}, pages
  8821--8831. PMLR.

\bibitem[{Ramnath et~al.(2021)Ramnath, Sari, Hasegawa-Johnson, and
  Yoo}]{ramnath-etal-2021-worldly}
Kiran Ramnath, Leda Sari, Mark Hasegawa-Johnson, and Chang Yoo. 2021.
\newblock \href {https://doi.org/10.18653/v1/2021.naacl-main.153} {Worldly wise
  ({W}o{W}) - cross-lingual knowledge fusion for fact-based visual
  spoken-question answering}.
\newblock In \emph{Proceedings of the 2021 Conference of the North American
  Chapter of the Association for Computational Linguistics: Human Language
  Technologies}, pages 1908--1919, Online. Association for Computational
  Linguistics.

\bibitem[{Ramos et~al.(2023)Ramos, Elliott, and
  Martins}]{ramos-etal-2023-retrieval}
Rita Ramos, Desmond Elliott, and Bruno Martins. 2023.
\newblock \href {https://aclanthology.org/2023.eacl-main.266}
  {Retrieval-augmented image captioning}.
\newblock In \emph{Proceedings of the 17th Conference of the European Chapter
  of the Association for Computational Linguistics}, pages 3666--3681,
  Dubrovnik, Croatia. Association for Computational Linguistics.

\bibitem[{Royal et~al.(2020)Royal, Hua, and Zhang}]{royal2020deep}
Brandon Royal, Kien Hua, and Brenton Zhang. 2020.
\newblock Deep composer: Deep neural hashing and retrieval approach to
  automatic music generation.
\newblock In \emph{2020 IEEE International Conference on Multimedia and Expo
  (ICME)}, pages 1--6. IEEE.

\bibitem[{Sarto et~al.(2022)Sarto, Cornia, Baraldi, and
  Cucchiara}]{RA-transformer}
Sara Sarto, Marcella Cornia, Lorenzo Baraldi, and Rita Cucchiara. 2022.
\newblock Retrieval-augmented transformer for image captioning.
\newblock In \emph{{CBMI}}, pages 1--7. {ACM}.

\bibitem[{Schick et~al.(2023)Schick, Dwivedi-Yu, Dess{\`\i}, Raileanu, Lomeli,
  Zettlemoyer, Cancedda, and Scialom}]{schick2023toolformer}
Timo Schick, Jane Dwivedi-Yu, Roberto Dess{\`\i}, Roberta Raileanu, Maria
  Lomeli, Luke Zettlemoyer, Nicola Cancedda, and Thomas Scialom. 2023.
\newblock Toolformer: Language models can teach themselves to use tools.
\newblock \emph{arXiv preprint arXiv:2302.04761}.

\bibitem[{Sciavolino et~al.(2021)Sciavolino, Zhong, Lee, and
  Chen}]{sciavolino-etal-2021-simple}
Christopher Sciavolino, Zexuan Zhong, Jinhyuk Lee, and Danqi Chen. 2021.
\newblock \href {https://doi.org/10.18653/v1/2021.emnlp-main.496} {Simple
  entity-centric questions challenge dense retrievers}.
\newblock In \emph{Proceedings of the 2021 Conference on Empirical Methods in
  Natural Language Processing}, pages 6138--6148, Online and Punta Cana,
  Dominican Republic. Association for Computational Linguistics.

\bibitem[{Selvaraju et~al.(2017)Selvaraju, Cogswell, Das, Vedantam, Parikh, and
  Batra}]{GradCAM}
Ramprasaath~R. Selvaraju, Michael Cogswell, Abhishek Das, Ramakrishna Vedantam,
  Devi Parikh, and Dhruv Batra. 2017.
\newblock Grad-cam: Visual explanations from deep networks via gradient-based
  localization.
\newblock In \emph{{ICCV}}, pages 618--626. {IEEE} Computer Society.

\bibitem[{Shen et~al.(2021)Shen, Zhan, Shen, Song, and
  Zhao}]{text_is_not_enough}
Lei Shen, Haolan Zhan, Xin Shen, Yonghao Song, and Xiaofang Zhao. 2021.
\newblock \href {https://doi.org/10.1145/3474085.3475568} {Text is not enough:
  Integrating visual impressions into open-domain dialogue generation}.
\newblock In \emph{Proceedings of the 29th ACM International Conference on
  Multimedia}, MM '21, page 4287–4296, New York, NY, USA. Association for
  Computing Machinery.

\bibitem[{Shi et~al.(2022)Shi, Wang, Tao, Du, Zhang, Han, Zhang, and
  Sun}]{shi-etal-2022-race}
Ensheng Shi, Yanlin Wang, Wei Tao, Lun Du, Hongyu Zhang, Shi Han, Dongmei
  Zhang, and Hongbin Sun. 2022.
\newblock \href {https://aclanthology.org/2022.emnlp-main.372} {{RACE}:
  Retrieval-augmented commit message generation}.
\newblock In \emph{Proceedings of the 2022 Conference on Empirical Methods in
  Natural Language Processing}, pages 5520--5530, Abu Dhabi, United Arab
  Emirates. Association for Computational Linguistics.

\bibitem[{Shi et~al.(2021)Shi, Liu, Min, Malon, Li, and
  Zhu}]{shi-etal-2021-retrieval-analogy}
Zhan Shi, Hui Liu, Martin~Renqiang Min, Christopher Malon, Li~Erran Li, and
  Xiaodan Zhu. 2021.
\newblock \href {https://doi.org/10.18653/v1/2021.findings-emnlp.171}
  {Retrieval, analogy, and composition: A framework for compositional
  generalization in image captioning}.
\newblock In \emph{Findings of the Association for Computational Linguistics:
  EMNLP 2021}, pages 1990--2000, Punta Cana, Dominican Republic. Association
  for Computational Linguistics.

\bibitem[{Shu et~al.(2022)Shu, Yu, Li, Karlsson, Ma, Qu, and
  Lin}]{shu-etal-2022-tiara}
Yiheng Shu, Zhiwei Yu, Yuhan Li, B{\"o}rje Karlsson, Tingting Ma, Yuzhong Qu,
  and Chin-Yew Lin. 2022.
\newblock \href {https://aclanthology.org/2022.emnlp-main.555} {{TIARA}:
  Multi-grained retrieval for robust question answering over large knowledge
  base}.
\newblock In \emph{Proceedings of the 2022 Conference on Empirical Methods in
  Natural Language Processing}, pages 8108--8121, Abu Dhabi, United Arab
  Emirates. Association for Computational Linguistics.

\bibitem[{Su et~al.(2021)Su, Meng, Baker, and
  Collier}]{su-etal-2021-shot-table}
Yixuan Su, Zaiqiao Meng, Simon Baker, and Nigel Collier. 2021.
\newblock \href {https://doi.org/10.18653/v1/2021.findings-emnlp.77} {Few-shot
  table-to-text generation with prototype memory}.
\newblock In \emph{Findings of the Association for Computational Linguistics:
  EMNLP 2021}, pages 910--917, Punta Cana, Dominican Republic. Association for
  Computational Linguistics.

\bibitem[{Sun et~al.(2019)Sun, Myers, Vondrick, Murphy, and
  Schmid}]{sun2019videobert}
Chen Sun, Austin Myers, Carl Vondrick, Kevin Murphy, and Cordelia Schmid. 2019.
\newblock Videobert: A joint model for video and language representation
  learning.
\newblock In \emph{Proceedings of the IEEE/CVF international conference on
  computer vision}, pages 7464--7473.

\bibitem[{Sun et~al.(2021)Sun, Hou, Wang, Zhang, and Wang}]{sun-etal-2021-d2s}
Edward Sun, Yufang Hou, Dakuo Wang, Yunfeng Zhang, and Nancy X.~R. Wang. 2021.
\newblock \href {https://doi.org/10.18653/v1/2021.naacl-main.111} {{D}2{S}:
  Document-to-slide generation via query-based text summarization}.
\newblock In \emph{Proceedings of the 2021 Conference of the North American
  Chapter of the Association for Computational Linguistics: Human Language
  Technologies}, pages 1405--1418, Online. Association for Computational
  Linguistics.

\bibitem[{Tan et~al.(2022)Tan, Gu, Tao, Ling, Xu, Hu, Geng, and
  Jiang}]{tan-etal-2022-tegtok}
Chao-Hong Tan, Jia-Chen Gu, Chongyang Tao, Zhen-Hua Ling, Can Xu, Huang Hu,
  Xiubo Geng, and Daxin Jiang. 2022.
\newblock \href {https://doi.org/10.18653/v1/2022.findings-acl.125}
  {{T}eg{T}ok: Augmenting text generation via task-specific and open-world
  knowledge}.
\newblock In \emph{Findings of the Association for Computational Linguistics:
  ACL 2022}, pages 1597--1609, Dublin, Ireland. Association for Computational
  Linguistics.

\bibitem[{Thoppilan et~al.(2022)Thoppilan, De~Freitas, Hall, Shazeer,
  Kulshreshtha, Cheng, Jin, Bos, Baker, Du et~al.}]{thoppilan2022lamda}
Romal Thoppilan, Daniel De~Freitas, Jamie Hall, Noam Shazeer, Apoorv
  Kulshreshtha, Heng-Tze Cheng, Alicia Jin, Taylor Bos, Leslie Baker, Yu~Du,
  et~al. 2022.
\newblock Lamda: Language models for dialog applications.
\newblock \emph{arXiv preprint arXiv:2201.08239}.

\bibitem[{Tiong et~al.(2022)Tiong, Li, Li, Savarese, and
  Hoi}]{tiong-etal-2022-plug}
Anthony Meng~Huat Tiong, Junnan Li, Boyang Li, Silvio Savarese, and Steven~C.H.
  Hoi. 2022.
\newblock \href {https://aclanthology.org/2022.findings-emnlp.67}
  {Plug-and-play {VQA}: Zero-shot {VQA} by conjoining large pretrained models
  with zero training}.
\newblock In \emph{Findings of the Association for Computational Linguistics:
  EMNLP 2022}, pages 951--967, Abu Dhabi, United Arab Emirates. Association for
  Computational Linguistics.

\bibitem[{Trivedi et~al.(2022)Trivedi, Balasubramanian, Khot, and
  Sabharwal}]{trivedi2022interleaving}
Harsh Trivedi, Niranjan Balasubramanian, Tushar Khot, and Ashish Sabharwal.
  2022.
\newblock Interleaving retrieval with chain-of-thought reasoning for
  knowledge-intensive multi-step questions.
\newblock \emph{arXiv preprint arXiv:2212.10509}.

\bibitem[{Tsai et~al.(2019)Tsai, Bai, Liang, Kolter, Morency, and
  Salakhutdinov}]{tsai2019multimodal}
Yao-Hung~Hubert Tsai, Shaojie Bai, Paul~Pu Liang, J~Zico Kolter, Louis-Philippe
  Morency, and Ruslan Salakhutdinov. 2019.
\newblock Multimodal transformer for unaligned multimodal language sequences.
\newblock In \emph{Proceedings of the conference. Association for Computational
  Linguistics. Meeting}, volume 2019, page 6558. NIH Public Access.

\bibitem[{Wang et~al.(2021{\natexlab{a}})Wang, Zhu, and Yang}]{wang2021t2vlad}
Xiaohan Wang, Linchao Zhu, and Yi~Yang. 2021{\natexlab{a}}.
\newblock \href {https://doi.org/10.1109/CVPR46437.2021.00504} {{T2VLAD:}
  global-local sequence alignment for text-video retrieval}.
\newblock In \emph{{IEEE} Conference on Computer Vision and Pattern
  Recognition, {CVPR} 2021, virtual, June 19-25, 2021}, pages 5079--5088.
  Computer Vision Foundation / {IEEE}.

\bibitem[{Wang et~al.(2021{\natexlab{b}})Wang, Wang, Joty, and
  Hoi}]{wang-etal-2021-codet5}
Yue Wang, Weishi Wang, Shafiq Joty, and Steven~C.H. Hoi. 2021{\natexlab{b}}.
\newblock \href {https://doi.org/10.18653/v1/2021.emnlp-main.685} {{C}ode{T}5:
  Identifier-aware unified pre-trained encoder-decoder models for code
  understanding and generation}.
\newblock In \emph{Proceedings of the 2021 Conference on Empirical Methods in
  Natural Language Processing}, pages 8696--8708, Online and Punta Cana,
  Dominican Republic. Association for Computational Linguistics.

\bibitem[{Wang et~al.(2022)Wang, Li, Xu, Zhou, Lei, Lin, Wang, Yang, Zhu, Hoiem
  et~al.}]{wang2022language}
Zhenhailong Wang, Manling Li, Ruochen Xu, Luowei Zhou, Jie Lei, Xudong Lin,
  Shuohang Wang, Ziyi Yang, Chenguang Zhu, Derek Hoiem, et~al. 2022.
\newblock Language models with image descriptors are strong few-shot
  video-language learners.
\newblock \emph{arXiv preprint arXiv:2205.10747}.

\bibitem[{Wei et~al.(2022)Wei, Wang, Schuurmans, Bosma, Chi, Le, and
  Zhou}]{wei2022chain}
Jason Wei, Xuezhi Wang, Dale Schuurmans, Maarten Bosma, Ed~Chi, Quoc Le, and
  Denny Zhou. 2022.
\newblock Chain of thought prompting elicits reasoning in large language
  models.
\newblock \emph{arXiv preprint arXiv:2201.11903}.

\bibitem[{Wenzel et~al.(2008)Wenzel, Paulson, and Nipkow}]{WenzelPN-TPHOLs08}
Makarius Wenzel, Lawrence~C. Paulson, and Tobias Nipkow. 2008.
\newblock The isabelle framework.
\newblock In \emph{TPHOLs}, volume 5170 of \emph{LNCS}, pages 33--38. Springer.

\bibitem[{Weston et~al.(2018)Weston, Dinan, and
  Miller}]{weston-etal-2018-retrieve}
Jason Weston, Emily Dinan, and Alexander Miller. 2018.
\newblock \href {https://doi.org/10.18653/v1/W18-5713} {Retrieve and refine:
  Improved sequence generation models for dialogue}.
\newblock In \emph{Proceedings of the 2018 {EMNLP} Workshop {SCAI}: The 2nd
  International Workshop on Search-Oriented Conversational {AI}}, pages 87--92,
  Brussels, Belgium. Association for Computational Linguistics.

\bibitem[{White et~al.(2019)White, Tufano, Martinez, Martin, and
  Poshyvanyk}]{White2019SortingAT}
Martin White, Michele Tufano, Matias Martinez, Monperrus Martin, and Denys
  Poshyvanyk. 2019.
\newblock Sorting and transforming program repair ingredients via deep learning
  code similarities.
\newblock \emph{2019 IEEE 26th International Conference on Software Analysis,
  Evolution and Reengineering (SANER)}, pages 479--490.

\bibitem[{Whitehead et~al.(2018)Whitehead, Ji, Bansal, Chang, and
  Voss}]{whitehead-etal-2018-incorporating}
Spencer Whitehead, Heng Ji, Mohit Bansal, Shih-Fu Chang, and Clare Voss. 2018.
\newblock \href {https://doi.org/10.18653/v1/D18-1433} {Incorporating
  background knowledge into video description generation}.
\newblock In \emph{Proceedings of the 2018 Conference on Empirical Methods in
  Natural Language Processing}, pages 3992--4001, Brussels, Belgium.
  Association for Computational Linguistics.

\bibitem[{Wu et~al.(2020{\natexlab{a}})Wu, Li, Zhang, and
  Wu}]{wu-etal-2020-improving-knowledge}
Sixing Wu, Ying Li, Dawei Zhang, and Zhonghai Wu. 2020{\natexlab{a}}.
\newblock \href {https://doi.org/10.18653/v1/2020.findings-emnlp.126}
  {Improving knowledge-aware dialogue response generation by using
  human-written prototype dialogues}.
\newblock In \emph{Findings of the Association for Computational Linguistics:
  EMNLP 2020}, pages 1402--1411, Online. Association for Computational
  Linguistics.

\bibitem[{Wu et~al.(2020{\natexlab{b}})Wu, Li, Zhang, Zhou, and
  Wu}]{wu-etal-2020-diverse}
Sixing Wu, Ying Li, Dawei Zhang, Yang Zhou, and Zhonghai Wu.
  2020{\natexlab{b}}.
\newblock \href {https://doi.org/10.18653/v1/2020.acl-main.515} {Diverse and
  informative dialogue generation with context-specific commonsense knowledge
  awareness}.
\newblock In \emph{Proceedings of the 58th Annual Meeting of the Association
  for Computational Linguistics}, pages 5811--5820, Online. Association for
  Computational Linguistics.

\bibitem[{Wu et~al.(2022{\natexlab{a}})Wu, Yang, Qiu, Ge, Yan, Wu, Zheng, Zhou,
  and Xiao}]{wu-etal-2022-deltanet}
Xian Wu, Shuxin Yang, Zhaopeng Qiu, Shen Ge, Yangtian Yan, Xingwang Wu, Yefeng
  Zheng, S.~Kevin Zhou, and Li~Xiao. 2022{\natexlab{a}}.
\newblock \href {https://aclanthology.org/2022.coling-1.261} {{D}elta{N}et:
  Conditional medical report generation for {COVID}-19 diagnosis}.
\newblock In \emph{Proceedings of the 29th International Conference on
  Computational Linguistics}, pages 2952--2961, Gyeongju, Republic of Korea.
  International Committee on Computational Linguistics.

\bibitem[{Wu et~al.(2019)Wu, Wei, Huang, Wang, Li, and Zhou}]{wu2019response}
Yu~Wu, Furu Wei, Shaohan Huang, Yunli Wang, Zhoujun Li, and Ming Zhou. 2019.
\newblock Response generation by context-aware prototype editing.
\newblock In \emph{AAAI}, volume~33, pages 7281--7288.

\bibitem[{Wu et~al.(2022{\natexlab{b}})Wu, Jiang, Li, Rabe, Staats, Jamnik, and
  Szegedy}]{wu2022autoformalization}
Yuhuai Wu, Albert~Qiaochu Jiang, Wenda Li, Markus~Norman Rabe, Charles~E
  Staats, Mateja Jamnik, and Christian Szegedy. 2022{\natexlab{b}}.
\newblock \href {https://openreview.net/forum?id=IUikebJ1Bf0}
  {Autoformalization with large language models}.
\newblock In \emph{NeurIPS}.

\bibitem[{Xia et~al.(2017)Xia, Bao, Lo, Kochhar, Hassan, and Xing}]{dsf}
Xin Xia, Lingfeng Bao, David Lo, Pavneet~Singh Kochhar, Ahmed~E. Hassan, and
  Zhenchang Xing. 2017.
\newblock \href {https://doi.org/10.1007/s10664-017-9514-4} {What do developers
  search for on the web?}
\newblock \emph{Empir. Softw. Eng.}, 22(6):3149--3185.

\bibitem[{Xin et~al.(2021)Xin, Hao, Dawei, and
  Yunfang}]{xin-etal-2021-enhancing}
Jia Xin, Wang Hao, Yin Dawei, and Wu~Yunfang. 2021.
\newblock \href {https://aclanthology.org/2021.ccl-1.87} {Enhancing question
  generation with commonsense knowledge}.
\newblock In \emph{Proceedings of the 20th Chinese National Conference on
  Computational Linguistics}, pages 976--987, Huhhot, China. Chinese
  Information Processing Society of China.

\bibitem[{Xu et~al.(2020)Xu, Crego, and Senellart}]{xu2020boosting}
Jitao Xu, Josep-Maria Crego, and Jean Senellart. 2020.
\newblock Boosting neural machine translation with similar translations.
\newblock In \emph{Annual Meeting of the Association for Computational
  Linguistics}, pages 1570--1579. Association for Computational Linguistics.

\bibitem[{Xu et~al.(2015)Xu, Xiong, Chen, and Corso}]{xu2015jointly}
Ran Xu, Caiming Xiong, Wei Chen, and Jason Corso. 2015.
\newblock Jointly modeling deep video and compositional text to bridge vision
  and language in a unified framework.
\newblock In \emph{AAAI}, volume~29.

\bibitem[{Xu et~al.(2021)Xu, Zhu, Xu, Liu, Zeng, and
  Huang}]{xu-etal-2021-fusing}
Yichong Xu, Chenguang Zhu, Ruochen Xu, Yang Liu, Michael Zeng, and Xuedong
  Huang. 2021.
\newblock \href {https://doi.org/10.18653/v1/2021.findings-acl.102} {Fusing
  context into knowledge graph for commonsense question answering}.
\newblock In \emph{Findings of the Association for Computational Linguistics:
  ACL-IJCNLP 2021}, pages 1201--1207, Online. Association for Computational
  Linguistics.

\bibitem[{Yamaguchi et~al.(2014)Yamaguchi, Golde, Arp, and
  Rieck}]{DBLP:conf/sp/YamaguchiGAR14}
Fabian Yamaguchi, Nico Golde, Daniel Arp, and Konrad Rieck. 2014.
\newblock \href {https://doi.org/10.1109/SP.2014.44} {Modeling and discovering
  vulnerabilities with code property graphs}.
\newblock In \emph{2014 {IEEE} Symposium on Security and Privacy, {SP} 2014,
  Berkeley, CA, USA, May 18-21, 2014}, pages 590--604. {IEEE} Computer Society.

\bibitem[{Yang et~al.(2023{\natexlab{a}})Yang, Nagrani, Seo, Miech,
  Pont{-}Tuset, Laptev, Sivic, and Schmid}]{yang23vid2seq}
Antoine Yang, Arsha Nagrani, Paul~Hongsuck Seo, Antoine Miech, Jordi
  Pont{-}Tuset, Ivan Laptev, Josef Sivic, and Cordelia Schmid.
  2023{\natexlab{a}}.
\newblock \href {https://doi.org/10.48550/arXiv.2302.14115} {Vid2seq:
  Large-scale pretraining of a visual language model for dense video
  captioning}.
\newblock \emph{CoRR}, abs/2302.14115.

\bibitem[{Yang et~al.(2021)Yang, Ye, You, and Ma}]{yang-etal-2021-writing}
Xingyi Yang, Muchao Ye, Quanzeng You, and Fenglong Ma. 2021.
\newblock \href {https://doi.org/10.18653/v1/2021.acl-long.387} {Writing by
  memorizing: Hierarchical retrieval-based medical report generation}.
\newblock In \emph{Proceedings of the 59th Annual Meeting of the Association
  for Computational Linguistics and the 11th International Joint Conference on
  Natural Language Processing (Volume 1: Long Papers)}, pages 5000--5009,
  Online. Association for Computational Linguistics.

\bibitem[{Yang et~al.(2022{\natexlab{a}})Yang, Yao, Zhang, Wang, Yu, and
  Chen}]{yang-etal-2022-z}
Yue Yang, Wenlin Yao, Hongming Zhang, Xiaoyang Wang, Dong Yu, and Jianshu Chen.
  2022{\natexlab{a}}.
\newblock \href {https://aclanthology.org/2022.emnlp-main.78} {{Z}-{L}a{VI}:
  Zero-shot language solver fueled by visual imagination}.
\newblock In \emph{Proceedings of the 2022 Conference on Empirical Methods in
  Natural Language Processing}, pages 1186--1203, Abu Dhabi, United Arab
  Emirates. Association for Computational Linguistics.

\bibitem[{Yang et~al.(2022{\natexlab{b}})Yang, Gan, Wang, Hu, Lu, Liu, and
  Wang}]{yang2022empirical}
Zhengyuan Yang, Zhe Gan, Jianfeng Wang, Xiaowei Hu, Yumao Lu, Zicheng Liu, and
  Lijuan Wang. 2022{\natexlab{b}}.
\newblock An empirical study of gpt-3 for few-shot knowledge-based vqa.
\newblock In \emph{AAAI}, volume~36, pages 3081--3089.

\bibitem[{Yang et~al.(2022{\natexlab{c}})Yang, Qin, Chen, Lin, and
  Liang}]{yang-etal-2022-logicsolver}
Zhicheng Yang, Jinghui Qin, Jiaqi Chen, Liang Lin, and Xiaodan Liang.
  2022{\natexlab{c}}.
\newblock \href {https://aclanthology.org/2022.findings-emnlp.1}
  {{L}ogic{S}olver: Towards interpretable math word problem solving with
  logical prompt-enhanced learning}.
\newblock In \emph{Findings of the Association for Computational Linguistics:
  EMNLP 2022}, pages 1--13, Abu Dhabi, United Arab Emirates. Association for
  Computational Linguistics.

\bibitem[{Yang et~al.(2023{\natexlab{b}})Yang, Ping, Liu, Korthikanti, Nie,
  Huang, Fan, Yu, Lan, Li, Liu, Zhu, Shoeybi, Catanzaro, Xiao, and
  Anandkumar}]{Re-ViLM}
Zhuolin Yang, Wei Ping, Zihan Liu, Vijay Korthikanti, Weili Nie, De{-}An Huang,
  Linxi Fan, Zhiding Yu, Shiyi Lan, Bo~Li, Ming{-}Yu Liu, Yuke Zhu, Mohammad
  Shoeybi, Bryan Catanzaro, Chaowei Xiao, and Anima Anandkumar.
  2023{\natexlab{b}}.
\newblock Re-vilm: Retrieval-augmented visual language model for zero and
  few-shot image captioning.
\newblock \emph{CoRR}, abs/2302.04858.

\bibitem[{Yasunaga et~al.(2022)Yasunaga, Aghajanyan, Shi, James, Leskovec,
  Liang, Lewis, Zettlemoyer, and Yih}]{RA-CM3}
Michihiro Yasunaga, Armen Aghajanyan, Weijia Shi, Rich James, Jure Leskovec,
  Percy Liang, Mike Lewis, Luke Zettlemoyer, and Wen{-}tau Yih. 2022.
\newblock Retrieval-augmented multimodal language modeling.
\newblock \emph{CoRR}, abs/2211.12561.

\bibitem[{Ye and Durrett(2022)}]{ye2022unreliability}
Xi~Ye and Greg Durrett. 2022.
\newblock \href {https://openreview.net/forum?id=Bct2f8fRd8S} {The
  unreliability of explanations in few-shot prompting for textual reasoning}.
\newblock In \emph{Advances in Neural Information Processing Systems}.

\bibitem[{Ye et~al.(2023)Ye, Hui, Yang, Li, Huang, and Li}]{ye2023large}
Yunhu Ye, Binyuan Hui, Min Yang, Binhua Li, Fei Huang, and Yongbin Li. 2023.
\newblock Large language models are versatile decomposers: Decompose evidence
  and questions for table-based reasoning.
\newblock \emph{arXiv preprint arXiv:2301.13808}.

\bibitem[{Yu et~al.(2022)Yu, Xu, Koh, Luong, Baid, Wang, Vasudevan, Ku, Yang,
  Ayan, Hutchinson, Han, Parekh, Li, Zhang, Baldridge, and Wu}]{parti}
Jiahui Yu, Yuanzhong Xu, Jing~Yu Koh, Thang Luong, Gunjan Baid, Zirui Wang,
  Vijay Vasudevan, Alexander Ku, Yinfei Yang, Burcu~Karagol Ayan, Ben
  Hutchinson, Wei Han, Zarana Parekh, Xin Li, Han Zhang, Jason Baldridge, and
  Yonghui Wu. 2022.
\newblock Scaling autoregressive models for content-rich text-to-image
  generation.
\newblock \emph{CoRR}, abs/2206.10789.

\bibitem[{Yu and Jiang(2021)}]{yu-jiang-2021-expanding}
Xiaojing Yu and Anxiao Jiang. 2021.
\newblock \href {https://doi.org/10.18653/v1/2021.eacl-main.279} {Expanding,
  retrieving and infilling: Diversifying cross-domain question generation with
  flexible templates}.
\newblock In \emph{Proceedings of the 16th Conference of the European Chapter
  of the Association for Computational Linguistics: Main Volume}, pages
  3202--3212, Online. Association for Computational Linguistics.

\bibitem[{Yuan et~al.(2023)Yuan, Jin, Tan, Zhao, Yuan, Huang, and Huang}]{RAMM}
Zheng Yuan, Qiao Jin, Chuanqi Tan, Zhengyun Zhao, Hongyi Yuan, Fei Huang, and
  Songfang Huang. 2023.
\newblock {RAMM:} retrieval-augmented biomedical visual question answering with
  multi-modal pre-training.
\newblock \emph{CoRR}, abs/2303.00534.

\bibitem[{Zeng et~al.(2022)Zeng, Attarian, Ichter, Choromanski, Wong, Welker,
  Tombari, Purohit, Ryoo, Sindhwani et~al.}]{zeng2022socratic}
Andy Zeng, Maria Attarian, Brian Ichter, Krzysztof Choromanski, Adrian Wong,
  Stefan Welker, Federico Tombari, Aveek Purohit, Michael Ryoo, Vikas
  Sindhwani, et~al. 2022.
\newblock Socratic models: Composing zero-shot multimodal reasoning with
  language.
\newblock \emph{arXiv preprint arXiv:2204.00598}.

\bibitem[{Zhang et~al.(2023{\natexlab{a}})Zhang, Li, and Bing}]{zhang2023video}
Hang Zhang, Xin Li, and Lidong Bing. 2023{\natexlab{a}}.
\newblock Video-llama: An instruction-tuned audio-visual language model for
  video understanding.
\newblock \emph{arXiv preprint arXiv:2306.02858}.

\bibitem[{Zhang et~al.(2020)Zhang, Wang, Zhang, Sun, and
  Liu}]{10.1145/3377811.3380383}
Jian Zhang, Xu~Wang, Hongyu Zhang, Hailong Sun, and Xudong Liu. 2020.
\newblock \href {https://doi.org/10.1145/3377811.3380383} {Retrieval-based
  neural source code summarization}.
\newblock In \emph{Proceedings of the ACM/IEEE 42nd International Conference on
  Software Engineering}, ICSE '20, page 1385–1397, New York, NY, USA.
  Association for Computing Machinery.

\bibitem[{Zhang et~al.(2018)Zhang, Utiyama, Sumita, Neubig, and
  Nakamura}]{zhang-etal-2018-guiding}
Jingyi Zhang, Masao Utiyama, Eiichro Sumita, Graham Neubig, and Satoshi
  Nakamura. 2018.
\newblock \href {https://doi.org/10.18653/v1/N18-1120} {Guiding neural machine
  translation with retrieved translation pieces}.
\newblock In \emph{Proceedings of the 2018 Conference of the North {A}merican
  Chapter of the Association for Computational Linguistics: Human Language
  Technologies, Volume 1 (Long Papers)}, pages 1325--1335, New Orleans,
  Louisiana. Association for Computational Linguistics.

\bibitem[{Zhang et~al.(2021)Zhang, Yang, Chen, He, and
  Yu}]{zhang-etal-2021-kers-knowledge}
Jun Zhang, Yan Yang, Chencai Chen, Liang He, and Zhou Yu. 2021.
\newblock \href {https://doi.org/10.18653/v1/2021.findings-emnlp.94} {{KERS}: A
  knowledge-enhanced framework for recommendation dialog systems with multiple
  subgoals}.
\newblock In \emph{Findings of the Association for Computational Linguistics:
  EMNLP 2021}, pages 1092--1101, Punta Cana, Dominican Republic. Association
  for Computational Linguistics.

\bibitem[{Zhang et~al.(2023{\natexlab{b}})Zhang, Zhang, Li, Zhao, Karypis, and
  Smola}]{zhang2023multimodal}
Zhuosheng Zhang, Aston Zhang, Mu~Li, Hai Zhao, George Karypis, and Alex Smola.
  2023{\natexlab{b}}.
\newblock Multimodal chain-of-thought reasoning in language models.
\newblock \emph{arXiv preprint arXiv:2302.00923}.

\bibitem[{Zhao et~al.(2023{\natexlab{a}})Zhao, Haffari, and
  Shareghi}]{zhao-etal-2023-generating}
Jinming Zhao, Gholamreza Haffari, and Ehsan Shareghi. 2023{\natexlab{a}}.
\newblock \href {https://aclanthology.org/2023.findings-eacl.147} {Generating
  synthetic speech from {S}poken{V}ocab for speech translation}.
\newblock In \emph{Findings of the Association for Computational Linguistics:
  EACL 2023}, pages 1975--1981, Dubrovnik, Croatia. Association for
  Computational Linguistics.

\bibitem[{Zhao et~al.(2023{\natexlab{b}})Zhao, Li, Chia, Ding, and
  Bing}]{zhao2023chatgptlike}
Ruochen Zhao, Xingxuan Li, Yew~Ken Chia, Bosheng Ding, and Lidong Bing.
  2023{\natexlab{b}}.
\newblock \href {http://arxiv.org/abs/2304.11076} {Can chatgpt-like generative
  models guarantee factual accuracy? on the mistakes of new generation search
  engines}.

\bibitem[{Zhao et~al.(2023{\natexlab{c}})Zhao, Li, Joty, Qin, and
  Bing}]{zhao2023verifyandedit}
Ruochen Zhao, Xingxuan Li, Shafiq Joty, Chengwei Qin, and Lidong Bing.
  2023{\natexlab{c}}.
\newblock \href {http://arxiv.org/abs/2305.03268} {Verify-and-edit: A
  knowledge-enhanced chain-of-thought framework}.

\bibitem[{Zhou et~al.(2020)Zhou, Palangi, Zhang, Hu, Corso, and
  Gao}]{zhou2020unified}
Luowei Zhou, Hamid Palangi, Lei Zhang, Houdong Hu, Jason Corso, and Jianfeng
  Gao. 2020.
\newblock Unified vision-language pre-training for image captioning and vqa.
\newblock In \emph{AAAI}, volume~34, pages 13041--13049.

\bibitem[{Zhou et~al.(2022{\natexlab{a}})Zhou, Luo, Rohrbach, and
  Yu}]{zhou-etal-2022-focus}
Mingyang Zhou, Grace Luo, Anna Rohrbach, and Zhou Yu. 2022{\natexlab{a}}.
\newblock \href {https://aclanthology.org/2022.findings-emnlp.450} {Focus!
  relevant and sufficient context selection for news image captioning}.
\newblock In \emph{Findings of the Association for Computational Linguistics:
  EMNLP 2022}, pages 6078--6088, Abu Dhabi, United Arab Emirates. Association
  for Computational Linguistics.

\bibitem[{Zhou et~al.(2022{\natexlab{b}})Zhou, Alon, Xu, Wang, Jiang, and
  Neubig}]{zhou2022docprompting}
Shuyan Zhou, Uri Alon, Frank~F Xu, Zhiruo Wang, Zhengbao Jiang, and Graham
  Neubig. 2022{\natexlab{b}}.
\newblock Docprompting: Generating code by retrieving the docs.
\newblock \emph{arXiv preprint arXiv:2207.05987}.

\bibitem[{Zhou and Long(2023)}]{zhou-long-2023-style}
Yucheng Zhou and Guodong Long. 2023.
\newblock \href {https://aclanthology.org/2023.findings-eacl.169} {Style-aware
  contrastive learning for multi-style image captioning}.
\newblock In \emph{Findings of the Association for Computational Linguistics:
  EACL 2023}, pages 2257--2267, Dubrovnik, Croatia. Association for
  Computational Linguistics.

\bibitem[{Zhu et~al.(2019)Zhu, Li, Zhu, Qian, Zhang, and
  Zhou}]{zhu-etal-2019-modeling}
Jie Zhu, Junhui Li, Muhua Zhu, Longhua Qian, Min Zhang, and Guodong Zhou. 2019.
\newblock \href {https://doi.org/10.18653/v1/D19-1548} {Modeling graph
  structure in transformer for better {AMR}-to-text generation}.
\newblock In \emph{Proceedings of the 2019 Conference on Empirical Methods in
  Natural Language Processing and the 9th International Joint Conference on
  Natural Language Processing (EMNLP-IJCNLP)}, pages 5459--5468, Hong Kong,
  China. Association for Computational Linguistics.

\bibitem[{Zhu et~al.(2023)Zhu, Yan, Lu, Xu, Wang, Eckstein, and
  Wang}]{zhu-etal-2023-visualize}
Wanrong Zhu, An~Yan, Yujie Lu, Wenda Xu, Xin Wang, Miguel Eckstein, and
  William~Yang Wang. 2023.
\newblock \href {https://aclanthology.org/2023.findings-eacl.5} {Visualize
  before you write: Imagination-guided open-ended text generation}.
\newblock In \emph{Findings of the Association for Computational Linguistics:
  EACL 2023}, pages 78--92, Dubrovnik, Croatia. Association for Computational
  Linguistics.

\end{thebibliography}
